\documentclass[12pt,a4paper]{article}
\usepackage{amssymb}
\usepackage{graphicx}
\usepackage{amsmath}
\usepackage{bbm}
\usepackage[sort&compress, square, numbers]{natbib}

\begin{document}

\title{Controlled Total Variation regularization \\
for inverse problems}
\author{Qiyu Jin$^{a,b}$ \and Ion Grama$^{a,b}$ \and Quansheng Liu$^{a,b}$ \\
\\
{\small {{qiyu.jin@univ-ubs.fr} \ \ \ \ \ \ \ {ion.grama@univ-ubs.fr}}}\\
{\small {{quansheng.liu@univ-ubs.fr} } }\\
{\small {$^{a}$Université de Bretagne-Sud, Campus de Tohaninic, BP 573,
56017 Vannes, France }}\\
{\small {\ $^{b}$Université Européenne de Bretagne, France } }}
\date{}
\maketitle

\begin{abstract}
This paper provides a new algorithm for solving inverse problems, based on
the minimization of the $L^2$ norm and on the control of the Total
Variation. It consists in relaxing the role of the Total Variation in the
classical Total Variation minimization approach, which permits us to get
better approximation to the inverse problems. The numerical results on the
deconvolution problem show that our method outperforms some previous ones.
\end{abstract}

\noindent

\section{Introduction}

Environmental effects and imperfections of image acquisition devices tend to
degrade the quality of imagery data, thereby making the problem of image
restoration an important part of modern imaging sciences. Very often the
images are distorted by some linear transformations. In this paper, we
consider the reconstruction of the unknown function $f$ from the following
inverse problem:
\begin{equation}
Y(x)=\mathcal{K}f(x) + \epsilon (x),\;x\in I=\left\{ 0,...,N-1\right\} ^{2},
\label{model convolution}
\end{equation}%
where $Y$ is the observed data, $\mathcal{K}$ is a continuous linear
operator from $R^{N^{2}}$ to $R^{M},$ $f$ is the original image, $\epsilon $
is a white Gaussian noise of mean $0$ and variance $\sigma ^{2},$ $N>1$ and $%
M>1$ are integers. We want to recover the original image $f$ starting from
the observed one $Y$. When the operator $\mathcal{K}$ is formulated as a
convolution with a kernel, this inverse problem reduces to the image
deconvolution model in the presence of noise. Of particular interest is the
case where the operator $\mathcal{K}$ is the convolution with the Gaussian
kernel:
\begin{equation}
\mathcal{K}f\left( x\right) =g\ast f\left( x\right) =\sum_{y\in I}g\left(
x-y\right) f\left( y\right) ,\;x\in I,  \label{K deconv}
\end{equation}%
where
\begin{equation}
g\left( y\right) =\frac{1}{2\pi \sigma _{b}^{2}}e^{-\frac{1}{2\sigma _{b}^{2}%
}\left\vert y\right\vert ^{2}},  \label{gaussian kernel}
\end{equation}%
and $\sigma _{b}>0$ is the standard deviation parameter. This type of
inverse problems appears in medical imaging, astronomical and laser imaging,
microscopy, remote sensing, photography, etc.

The problem (\ref{model convolution}) can be decomposed into two
sub-problems: the denoising problem%
\begin{equation}
Y(x)=u(x) + \epsilon (x),\;x\in I  \label{denoising}
\end{equation}%
and the possibly ill-posed inverse problem
\begin{equation}
u\left( x\right) =\mathcal{K}f\left( x\right) ,\;x\in I.
\label{model deconvolution}
\end{equation}%
Thus, the restoration of the signal $f$ from the model (\ref{model
convolution}) will be performed into two steps: firstly, remove the Gaussian
noise in the model (\ref{denoising}); secondly, find a solution to the
problem (\ref{model deconvolution}).

For the first step, there are many efficient methods, see e.g. Buades, Coll
and Morel (2005 \citep{Bu}), Kervrann (2006 \citep{kervrann2006optimal}),
Lou, Zhang, Osher and Bertozzi (2010 \citep{lou2010image}), Polzehl and
Spokoiny (2006 \citep{polzehl2006propagation}), Garnett, Huegerich and Chui
(2005 \citep{Garnett2005universal}), Cai, Chan, Nikolova (2008 %
\citep{cai2008two}), Katkovnik, Foi, Egiazarian, and Astola ( 2010 %
\citep{Katkovnik2010local}), Dabov, Foi, Katkovnik and Egiazarian (2006 %
\citep{buades2009note}), and Jin, Grama and Liu(2011 \citep{JinGramaLiuowf}).

The second step consists in finding a convenient function $\widetilde{f}$
satisfying (exactly or approximately)
\begin{equation}
\mathcal{K}\widetilde{f}-u=\mathbf{0},  \label{deconv approx}
\end{equation}%
where $u$ is obtained by the denoising algorithm of the first step. To find
a solution to the ill-posed problem (\ref{deconv approx}), one usually
considers some constraints which are believed to be satisfied by natural
images. Such constraints may include, in particular, the minimization of the
$\mathbf{L}^{2}$ norm of $\widetilde{f}$ or $\nabla \widetilde{f}$ and the
minimization of the Total Variation.

The famous Tikhonov regularization method (1977 \citep{Tikhonov1977solution}
and 1996 \citep{engl1996regularization}) consists in solving the following
minimization problem:
\begin{equation}
\widehat{f}(x)=\arg \min_{\widetilde{f}(x)}\;\Vert \widetilde{f}\Vert
_{2}^{2}  \label{tikhonov min}
\end{equation}%
subject to
\begin{equation}
\Vert \mathcal{K}\widetilde{f}-u\Vert _{2}^{2}=0.
\label{convolution condition}
\end{equation}%
Using the usual Lagrange multipliers approach, this leads to
\begin{equation*}
\widehat{f}=\arg \min_{\widetilde{f}}\;\Vert \widetilde{f}\Vert _{2}^{2} + %
\frac{1}{2}\lambda \Vert \mathcal{K}\widetilde{f}-u\Vert _{2}^{2},
\end{equation*}%
where $\lambda >0$ is interpreted as a regularization parameter.

An alternative regularization (1977 \citep{andrews1977digital}) is to
replace the $\mathbf{L}^{2}$ norm in (\ref{tikhonov min}) with the $\mathbf{H%
}^{1}$ semi-norm, which leads to the well-known Wiener filter for image
deconvolution:%
\begin{equation}
\widehat{f}=\arg \min_{\widetilde{f}}\;\Vert \nabla \widetilde{f}\Vert
_{2}^{2}  \label{H1 semi-norm}
\end{equation}%
subject to (\ref{convolution condition}), or, again by the Lagrange
multipliers approach,%
\begin{equation}
\widehat{f}=\arg \min_{\widetilde{f}}\;\Vert \nabla \widetilde{f}\Vert
_{2}^{2} + \frac{1}{2}\lambda \Vert \mathcal{K}\widetilde{f}-u\Vert _{2}^{2}.
\label{model semi-norm}
\end{equation}%
where $\lambda >0$ is a regularization parameter.

The Total Variation (TV) regularization called ROF model (see Rudin, Osher
and Fatemi \citep{rudin1992nonlinear}), is a modification of (\ref{H1
semi-norm}) with $\Vert \nabla \widetilde{f}\Vert _{2}^{2}$ replaced by $%
\Vert \nabla \widetilde{f}\Vert _{1}:$
\begin{equation}
\widehat{f}=\arg \min_{\widetilde{f}}\;\Vert \nabla \widetilde{f}\Vert _{1}
\label{model TV}
\end{equation}%
subject to (\ref{convolution condition}), or, again by the Lagrange
multipliers approach,%
\begin{equation}
\widehat{f}=\arg \min_{\widetilde{f}}\;\Vert \nabla \widetilde{f}\Vert _{1} + %
\frac{1}{2}\lambda \Vert \mathcal{K}\widetilde{f}-u\Vert _{2}^{2}.
\label{model TV2}
\end{equation}%
where $\lambda >0$ is a regularization parameter.

Due to its virtue of preserving edges, it is widely used in image
processing, such as blind deconvolution %
\citep{chan1998total,he2005blind,marquina2009nonlinear}, inpaintings %
\citep{chan2001mathematical} and super-resolution \citep{marquina2008image}.
Luo et al \citep{lou2010image} composed TV-based methods with Non-Local
means approach to preserve fine structures, details and textures.

The classical approach to resolve (\ref{model TV2}), for a fixed parameter $%
0<\lambda < + \infty ,$ consists in searching for a critical point $\widetilde{%
f}$ characterized by
\begin{equation}
L(\widetilde{f}) + \lambda \mathcal{K}^{\ast }(\mathcal{K}\widetilde{f}-u)=0,
\label{critique}
\end{equation}%
where $L(f)=\nabla \frac{\nabla {f}}{|\nabla {f}|},$ $|\nabla {f}|$ being
the Euclidean norm of the gradient $\nabla {f.}$ The usual technique to find
a solution $\widetilde{f}$ of (\ref{critique}) is the gradient descent
approach, which employs the iteration procedure
\begin{equation}
f_{k + 1}=f_{k}-h\left( Lf_{k}  + \lambda \mathcal{K}^{\ast }(\mathcal{K}%
f_{k}-u)\right) ,  \label{iterations}
\end{equation}%
where $h>0$ is a fixed parameter. If the algorithm converges, say $%
f_{k}\rightarrow \widetilde{f},$ then $\widetilde{f}$ satisfies the equation
(\ref{critique}). However, this algorithm cannot resolve exactly the initial
deconvolution problem (\ref{deconv approx}), for, usually, $L(\widetilde{f}%
)\neq 0,$ so that, by (\ref{critique}), $\mathcal{K}\widetilde{f}-u\neq 0.$
This will be shown in the next sections by simulation results.

In this paper, we propose an improvement of the algorithm (\ref{iterations}%
), which permits us to obtain an exact solution of the deconvolution problem
(\ref{deconv approx}) when the solution exists, and the closest
approximation to the original image when the exact solution to (\ref{deconv
approx}) does not exist.

 Precisely, we shall search for a solution $%
\widetilde{f}$ of the equation%
\begin{equation*}
\mathcal{K}^{\ast }(\mathcal{K}\widetilde{f}-u)=0,
\end{equation*}%
 with a small enough TV.
Notice that the last equation characterizes the critical points $\widetilde{f}$
of the minimization problem $\min_{\widetilde{f}}~\Vert \mathcal{K}%
\widetilde{f}-u\Vert _{2}^{2}$ which, in turn, whose minimiser gives the closest
approximation to the deconvolution problem (\ref{deconv approx}), even if it
does not have any solution. In the proposed algorithm, we do not search for
the minimization of the TV. Instead of this, we introduce a factor in the
iteration process (\ref{iterations}) to have a control on the evolution of
the TV. Compared with the usual TV minimization approach, the new algorithm
retains the image details with higher fidelity, as the TV of the restored image
is closer to the TV of the original image.

Our experimental results confirm this point, and show that our new algorithm
outperforms the recently ones proposed in the literature that are based on
the minimization problems (\ref{model semi-norm}) and (\ref{model TV2}).

\section{A\ new algorithm for the inverse problem}

\subsection{Controlled Total Variation regularization algorithm}

We would like to find $\widetilde{f}$ satisfying the deconvolution equation (%
\ref{deconv approx}) with a \textit{small Total Variation}. Since an exact
solution to (\ref{deconv approx}) may not exist, instead of the initial
problem we consider the minimization problem
\begin{equation}
\min_{\widetilde{f}}~\Vert \mathcal{K}\widetilde{f}-u\Vert _{2}^{2}.
\label{minimiz}
\end{equation}%
This leads us to search for critical points $\widetilde{f}$ characterized by
the equation%
\begin{equation}
\mathcal{K}^{\ast }(\mathcal{K}\widetilde{f}-u)=0,  \label{crit L2}
\end{equation}%
whose TV is small enough. As explained in the introduction, the usual
algorithm (\ref{iterations}) (classic TV) does not lead to a solution of (\ref{crit L2}) if
it exists. In fact, we have observed by simulation results that usually $%
\Vert Lf_{k}\Vert _{1}$ does not converge to $0$ with $k\rightarrow  + \infty ,
$ so that $\mathcal{K}^{\ast }(\mathcal{K}f_{k}-u)$ does not tend to $0$
neither.

\begin{figure}[tbp]
\begin{center}
\renewcommand{\arraystretch}{0.5} \addtolength{\tabcolsep}{-6pt} \vskip3mm {%
\fontsize{8pt}{\baselineskip}\selectfont
\begin{tabular}{cc}
\multicolumn{2}{c}{ Classical  TV and CTV (Gaussian blur kernel)}\\
\includegraphics[width=0.49\linewidth]{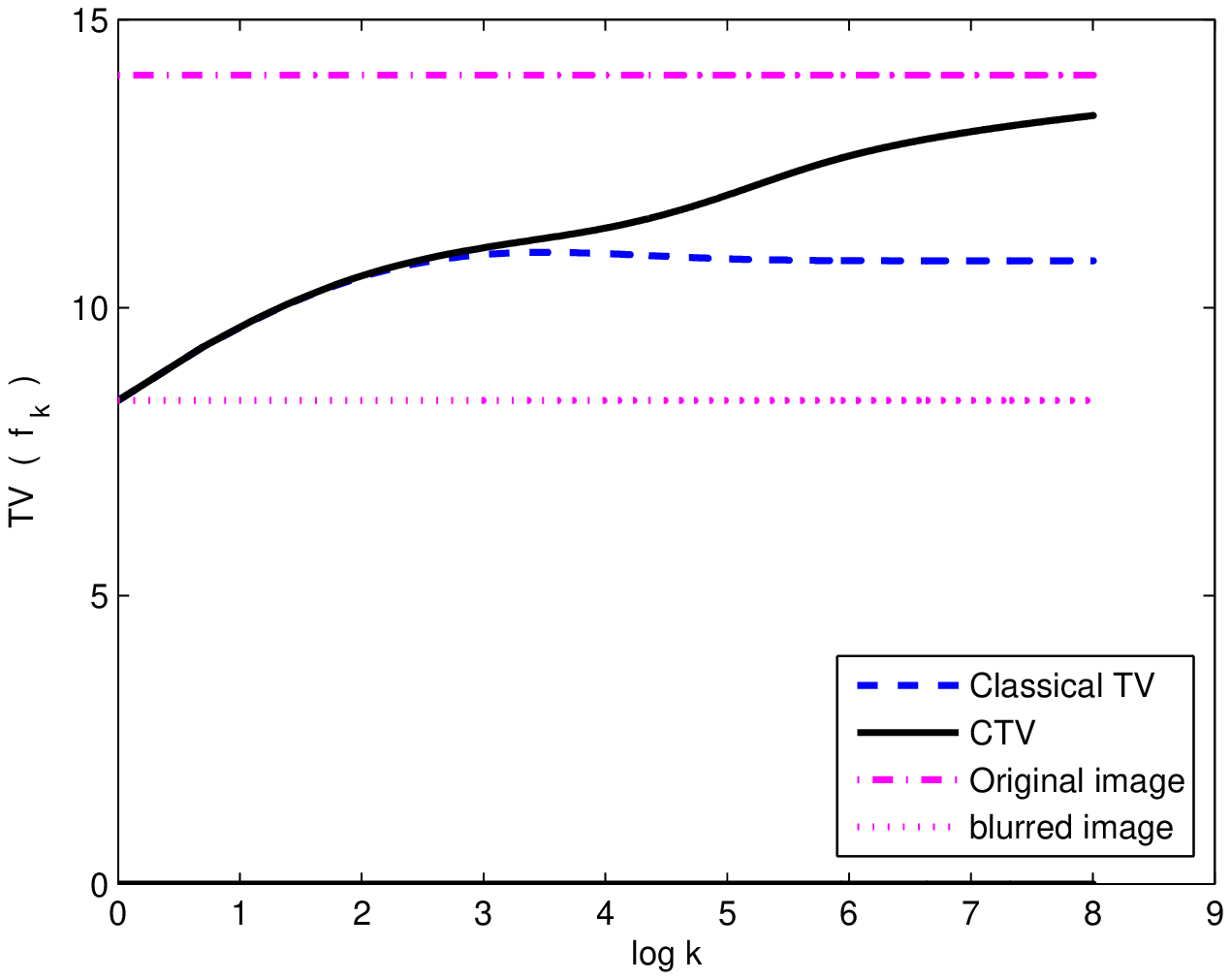} & %
\includegraphics[width=0.49\linewidth]{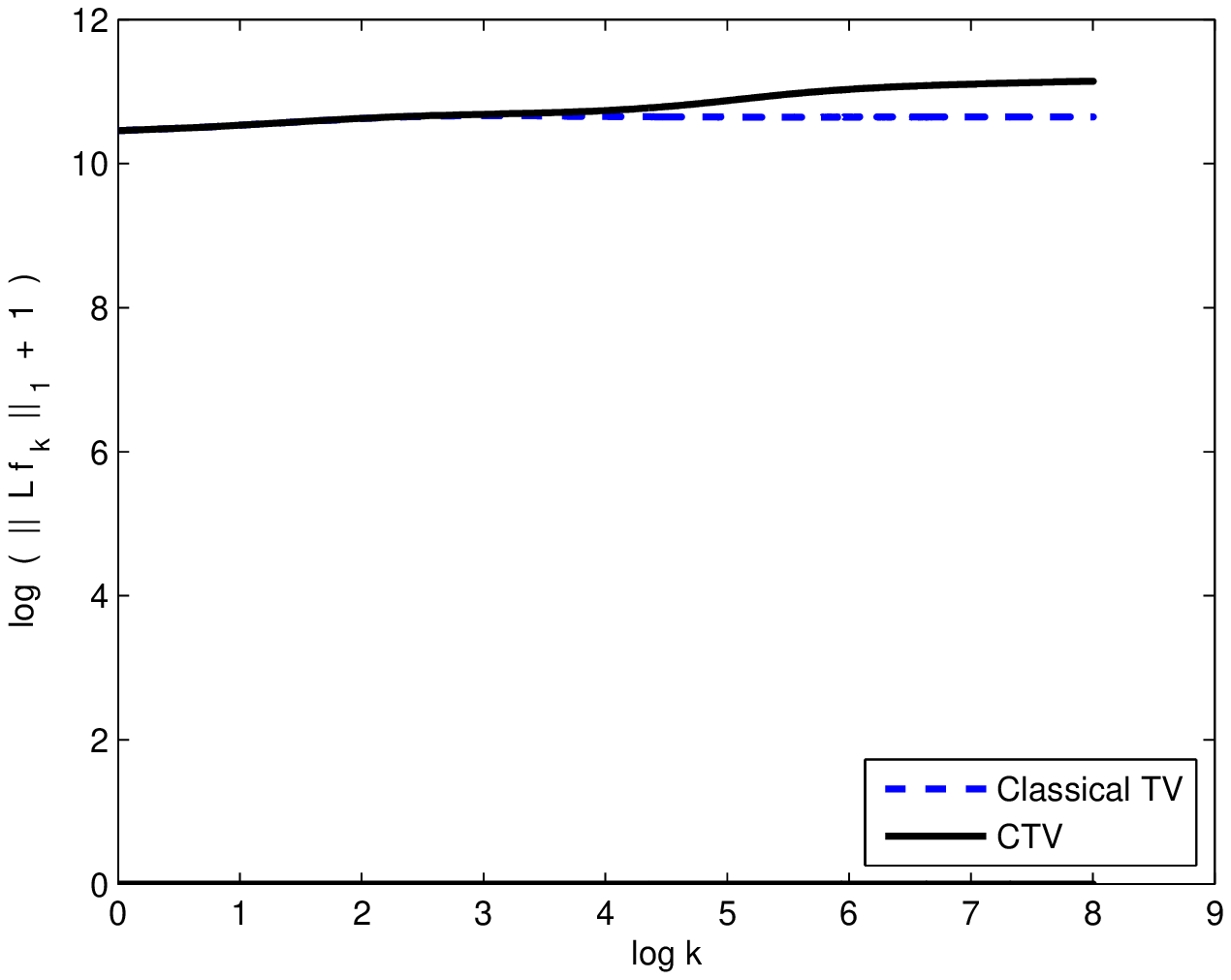} \\
(a) & (b) \\
\includegraphics[width=0.49\linewidth]{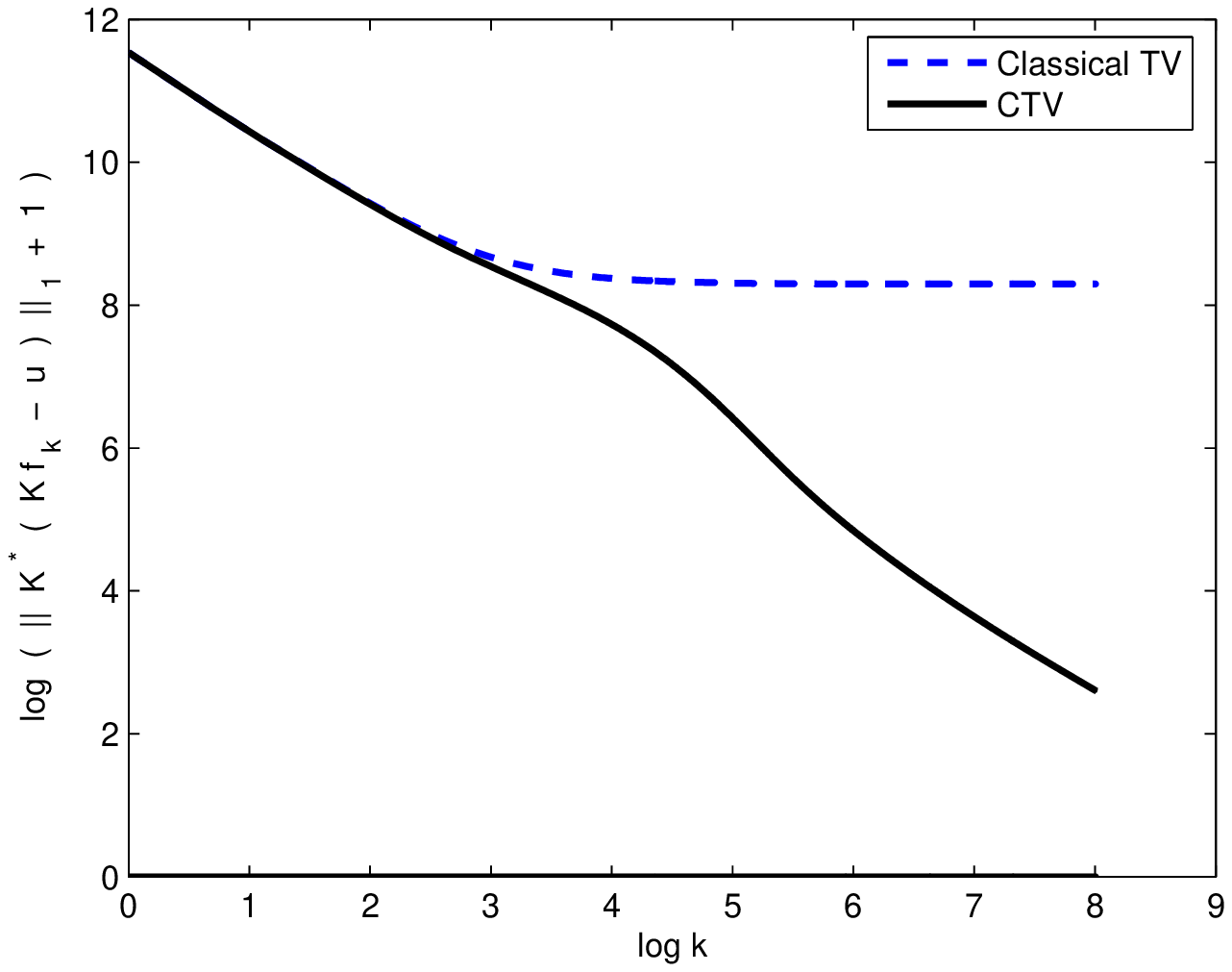} & %
\includegraphics[width=0.49\linewidth]{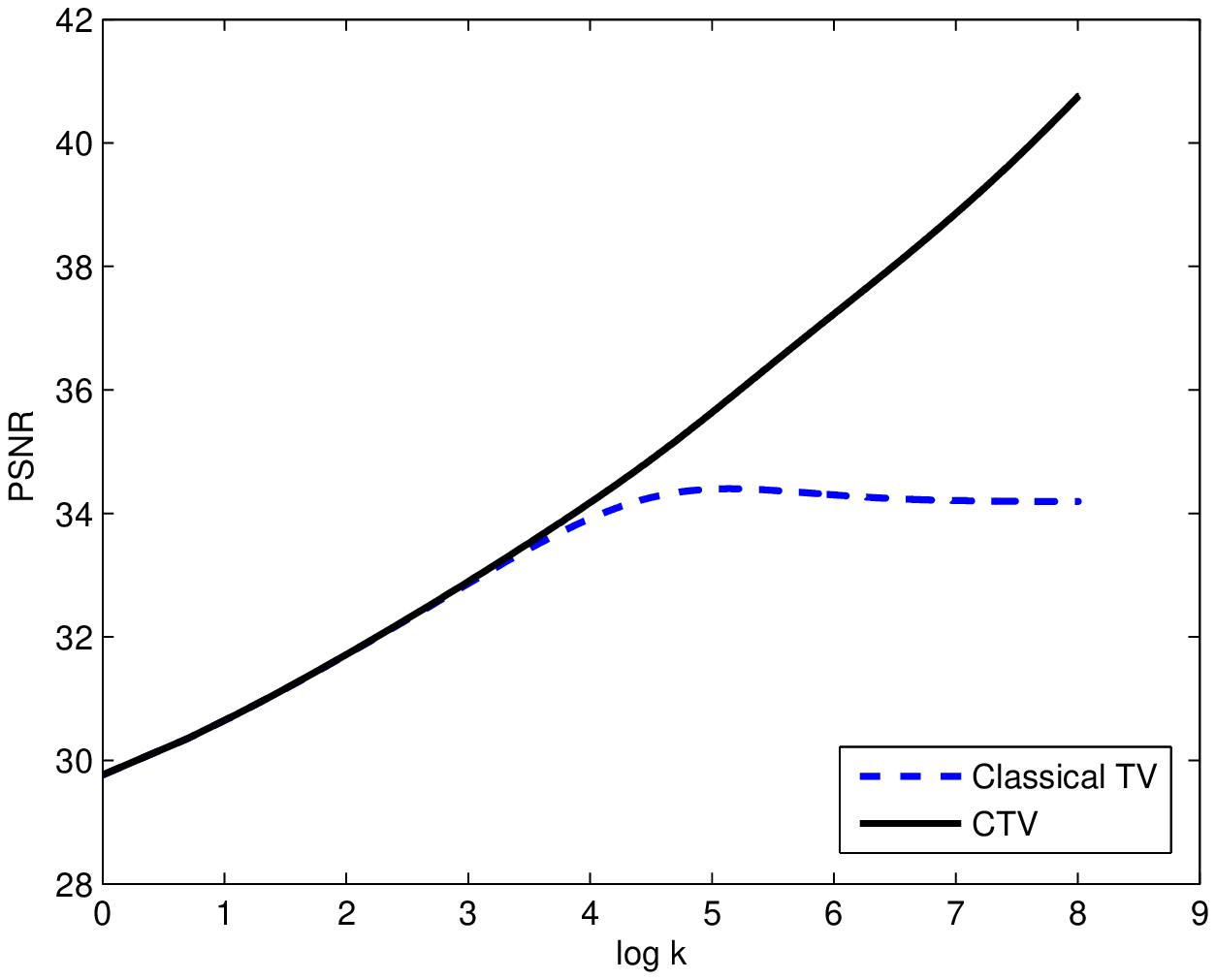} \\
(c) & (d)%
\end{tabular}
}
\end{center}
\caption{{\protect\small Comparison between classical TV and CTV. The test image "Lena" is convoluted with Gaussian
blur kernel of standard deviation $\protect\sigma_b=1$. We choose $h =0.1$, $%
\protect\lambda=1$, $\max Iter=3000$ and $\protect\theta=0.98$. (a) The
evolution of $TV(f_k)$ as a function of $\log k$. (b) The evolution of $%
\log(||Lf_k||_1 + 1)$ as a function of $\log k$. (c) The evolution of $\log(||%
\mathcal{K}^*(\mathcal{K}f_k-u)||_1 + 1)$ as a function of $\log k$, (d) The
evolution of PSNR value as a function of $\log k$.}}
\label{Fig gau etv}
\end{figure}

For example taking "Lena" as test image blurred by the Gaussian convolution
kernel (\ref{gaussian kernel}) with standard deviation $\sigma _{b}=2,$ the
value of $||Lf_{k}||_{1}$ increases (see Figure \ref{Fig gau etv} (b), where the classical TV curve correspond to the dashed line) as $%
\log k$ increases. On the other hand, Figure \ref{Fig gau etv} (c) shows
that the value of $||\mathcal{K}^{\ast }(\mathcal{K}f_{k}-u)||_{1}$
decreases when $\log k$ increases in the interval $[0,3]$, but it remains
bounded from below by $e^{8}-1\approx 2980$ and does not decrease towards $0$
when $\log k$ varies in the interval $[3,8]$.

These considerations lead us to the following modified version of (\ref%
{iterations}):
\begin{equation}
f_{k + 1}=f_{k}-h\left( \theta _{k} Lf_{k}  + \lambda \mathcal{K}^{\ast }(%
\mathcal{K}f_{k}-u)\right) ,  \label{iteration jin}
\end{equation}%
where $h>0,$ $\lambda >0$ are as in (\ref{iterations}) and $\theta _{k}>0$
is a sequence decreasing to $0.$ We shall use $\theta _{k}=\theta ^{k},$ for
some $\theta \in (0,1).$ This ensures that, if the algorithm converges, say $%
f_{k}\rightarrow \widetilde{f}$, then
\begin{equation}
\mathcal{K}^{\ast }(\mathcal{K}\widetilde{f}-u)=0.
\label{iteration energy jin}
\end{equation}

Notice that, compared with the initial algorithm (\ref{iterations}), we put
a small weight $\theta _{k}$ to the TV in the minimization problem (\ref%
{model TV2}), thus diminishing the role of this term in the minimization, in
order to get an exact solution to the ill-posed problem (\ref{iteration
energy jin}). Here we do not search for the minimization of the TV, but just for a control
on the TV. The control of TV rather than the minimization
of TV permets us to better preserve image edges and details.

The new algorithm given by (\ref{iteration jin}) will be called \textit{%
Controlled Total Variation } regularization algorithm (CTV).

\begin{figure}[tbp]
\begin{center}
\renewcommand{\arraystretch}{0.5} \addtolength{\tabcolsep}{-6pt} \vskip3mm {%
\fontsize{8pt}{\baselineskip}\selectfont
\begin{tabular}{cc}
\multicolumn{2}{c}{ Classical  TV and CTV (box average blue kernel)}\\
\includegraphics[width=0.49\linewidth]{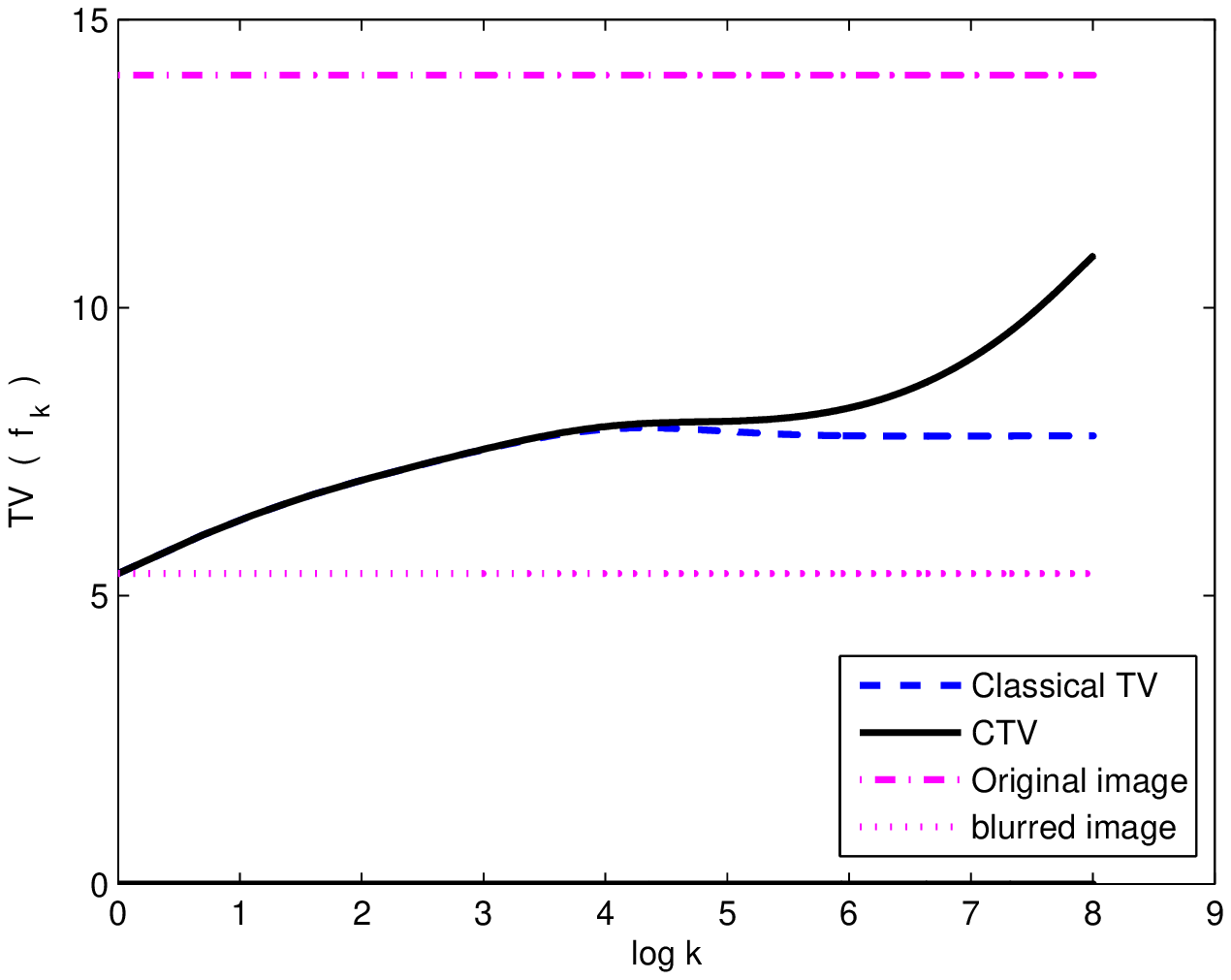} & %
\includegraphics[width=0.49\linewidth]{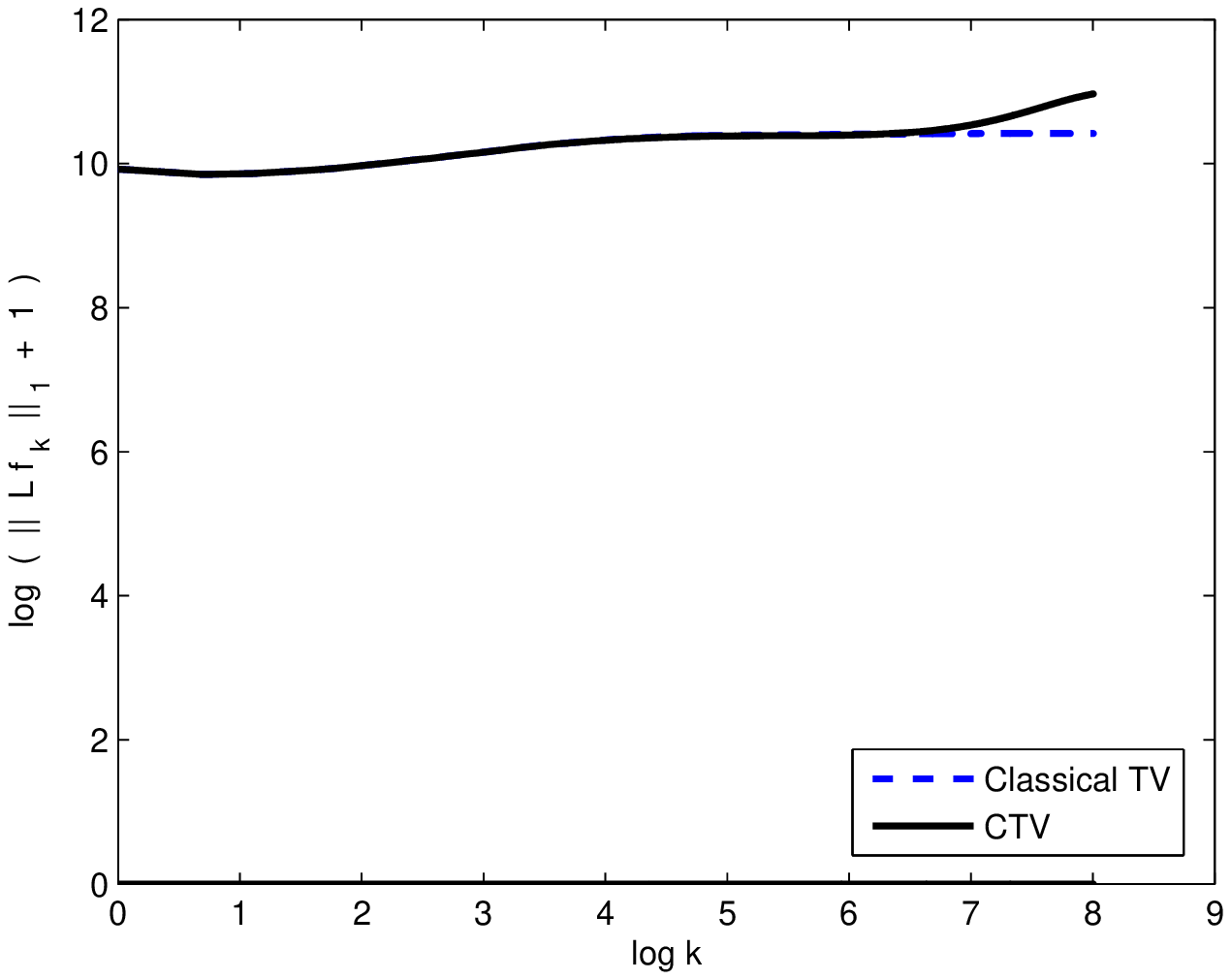} \\
(a) & (b) \\
\includegraphics[width=0.49\linewidth]{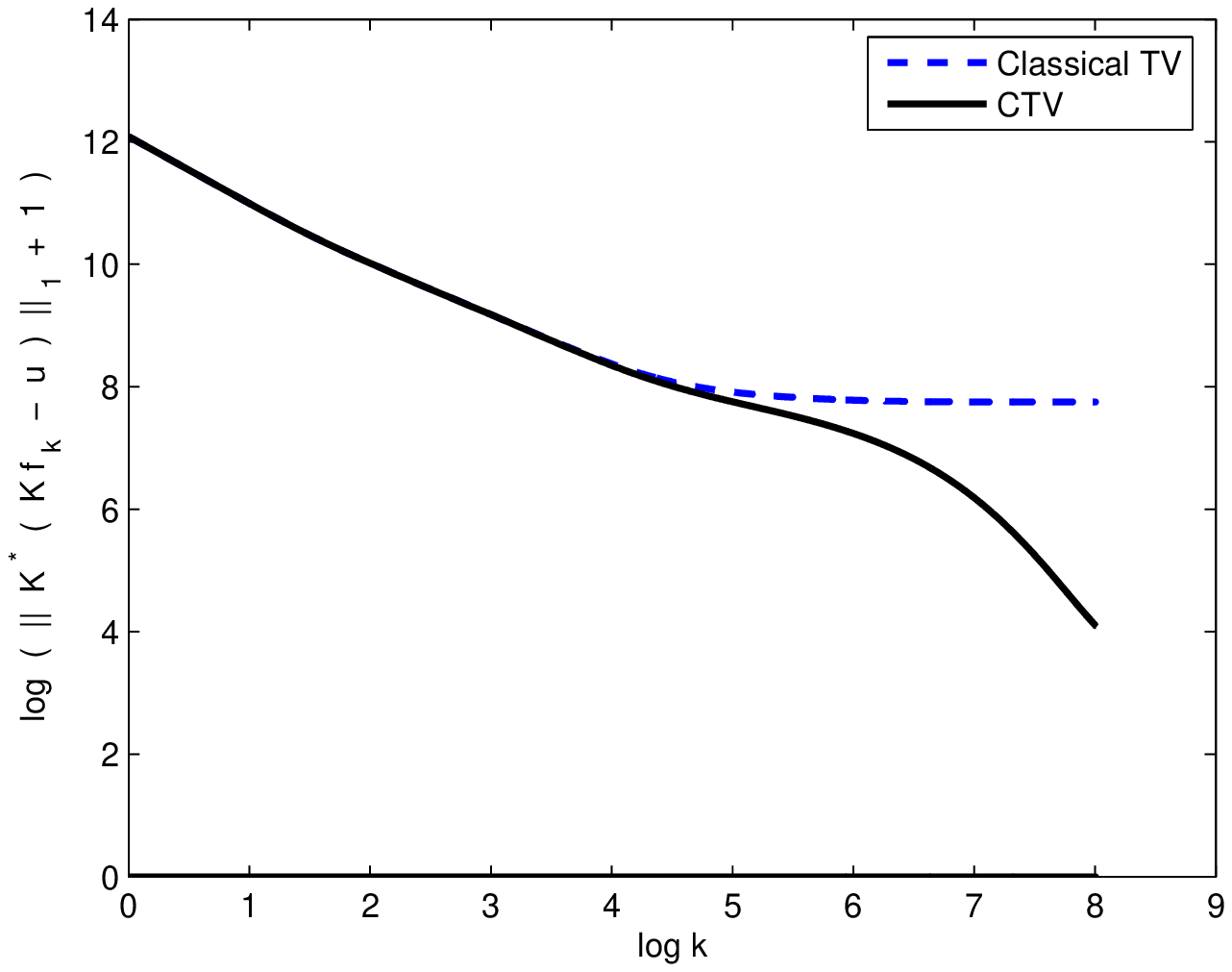} & %
\includegraphics[width=0.49\linewidth]{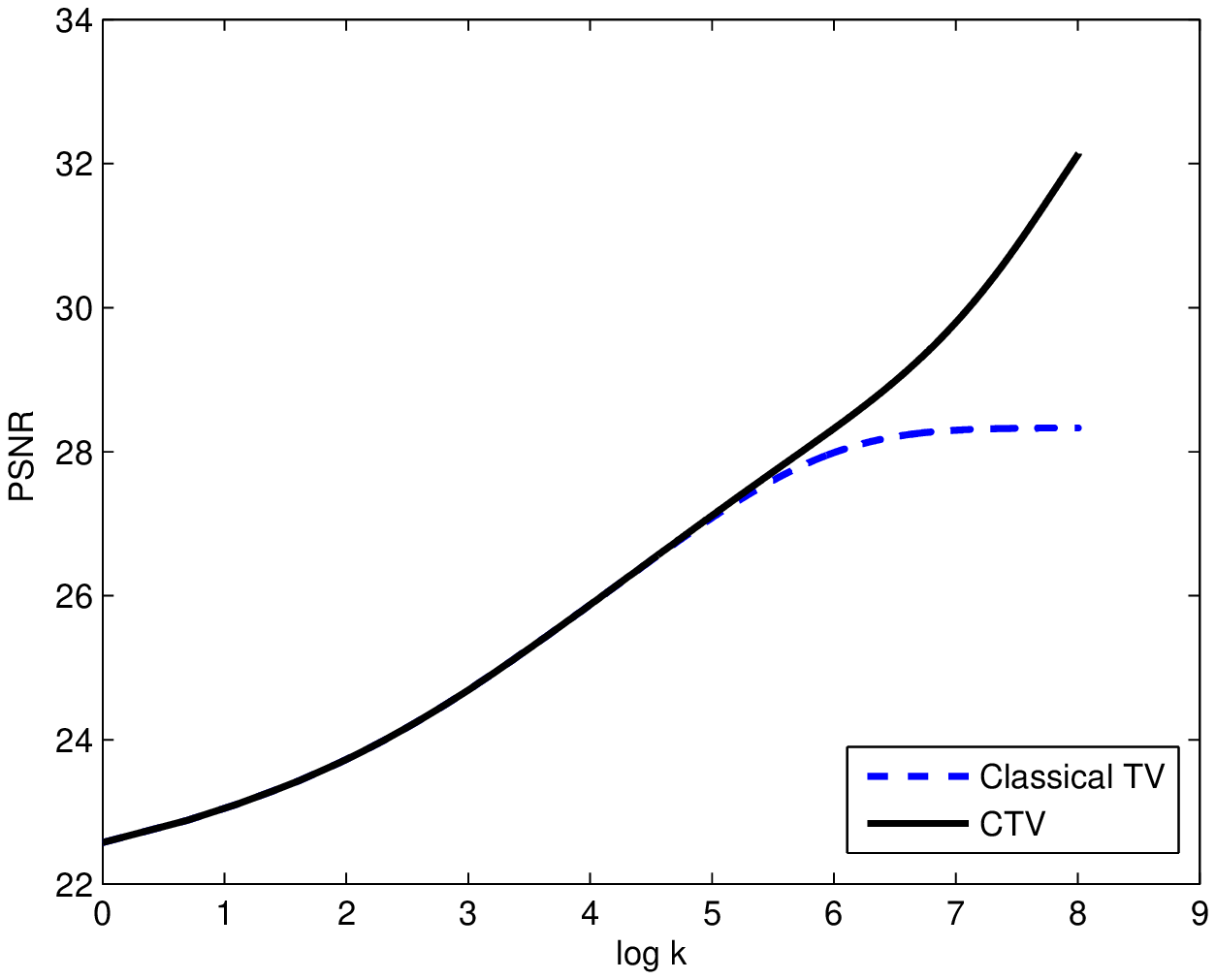} \\
(c) & (d)%
\end{tabular}
}
\end{center}
\caption{{\protect\small Comparison between classical TV and CTV.  The test image "Lena" is convoluted with $9\times9$
box average blur kernel. We choose $h =0.1$, $\protect\lambda=1$, $\max
Iter=3000$ and $\protect\theta=0.998$. (a) The evolution of $TV(f_k)$ as a
function of $\log k$. (b) The evolution of $\log(||Lf_k||_1 + 1)$ as a
function of $\log k$. (c) The evolution of $\log(||\mathcal{K}^*(\mathcal{K}%
f_k-u)||_1 + 1)$ as a function of $\log k$, (d) The evolution of PSNR value as
a function of $\log k$.}}
\label{Fig box etv}
\end{figure}

We show by
simulations that the new algorithm improves the quality of image
restauration.
In Figures \ref{Fig
gau etv} and \ref{Fig box etv}, we see that the TV of the restored image by
the proposed CTV algorithm is larger than that of the restored image by the
classical TV minimization approach, but closer to the TV of the original
image. This also shows that in the classical TV minimization algorithm the
TV of the restored image is too small and that the classical approach may
not give the best solution.

On the other hand, by simulation we have also seen that the TV in the
iterative process (\ref{iteration jin}) cannot be suppressed completely,
i.e. we cannot take $\theta _{k}=0.$

\begin{figure}[tbp]
\begin{center}
\renewcommand{\arraystretch}{0.5} \addtolength{\tabcolsep}{-6pt} \vskip3mm {%
\fontsize{8pt}{\baselineskip}\selectfont
\begin{tabular}{cc}
\multicolumn{2}{c}{ Restoration results with classical TV and CTV }\\
\includegraphics[width=0.49\linewidth]{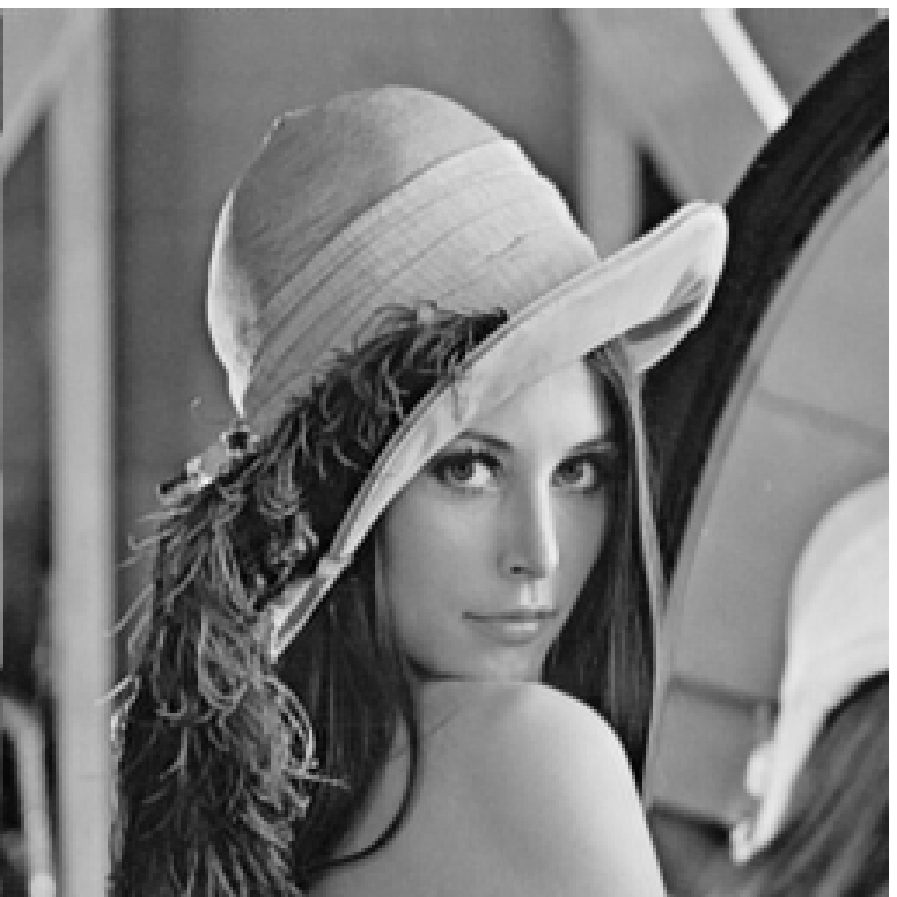} & %
\includegraphics[width=0.49\linewidth]{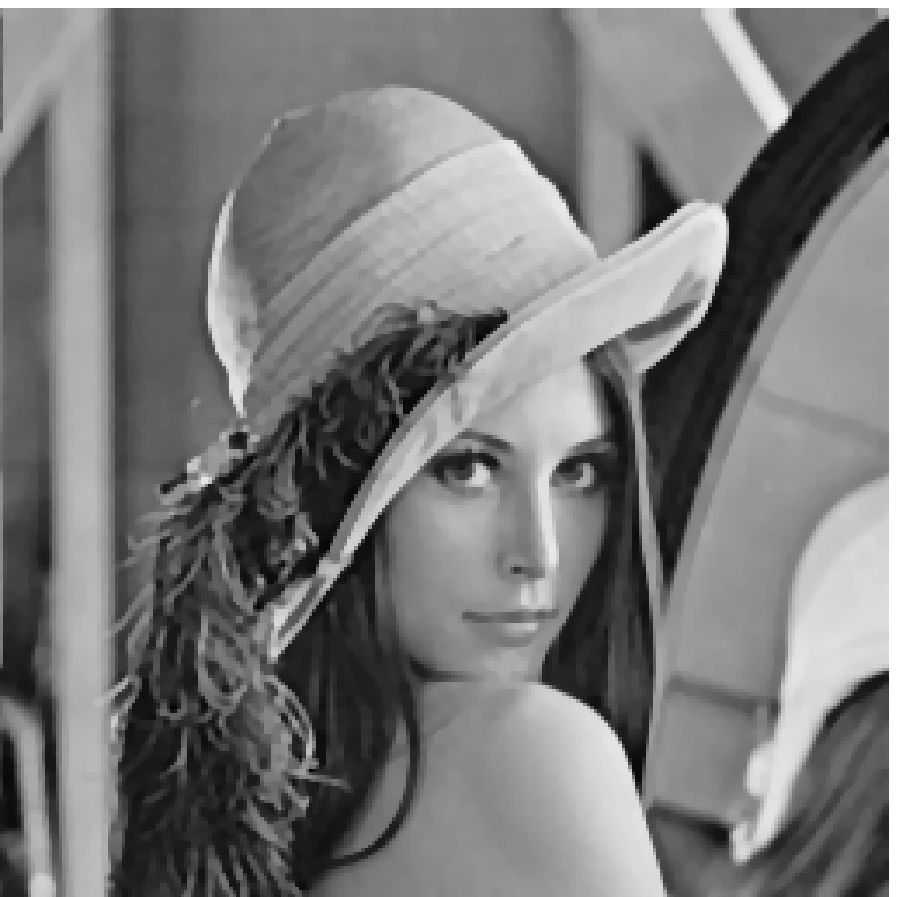} \\
(a) CTV, PSNR$=42.05$db      & (b) classical TV, PSNR$=34.18$db\\
\includegraphics[width=0.49\linewidth]{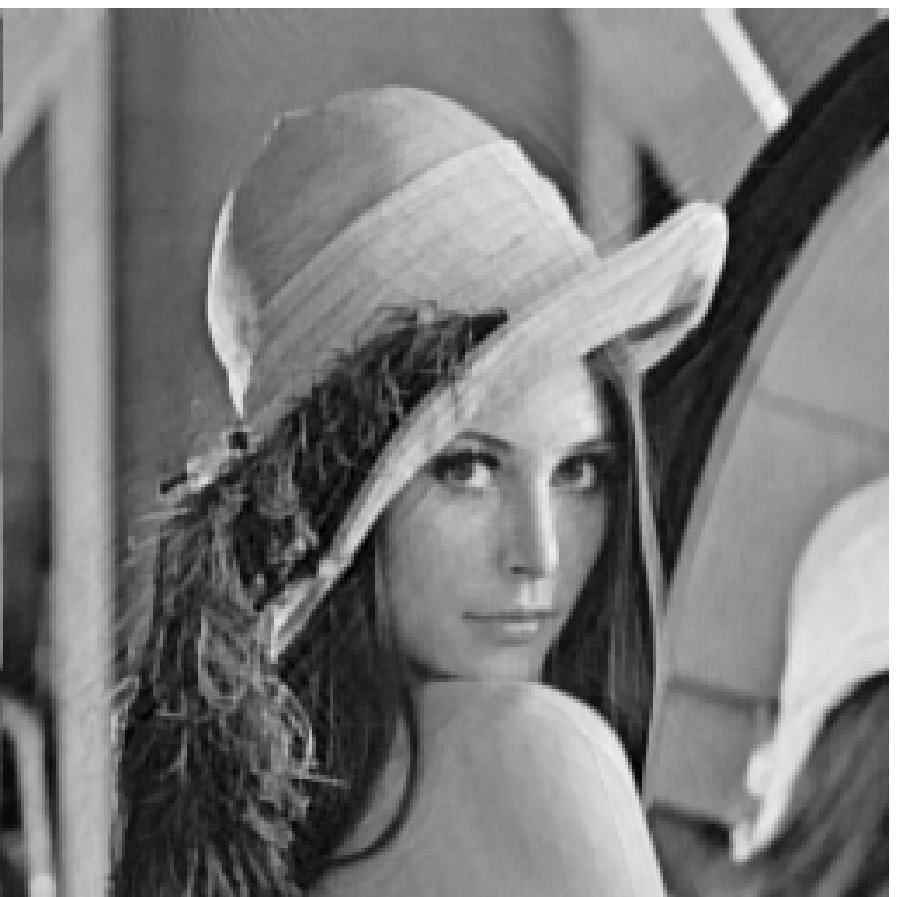} & %
\includegraphics[width=0.49\linewidth]{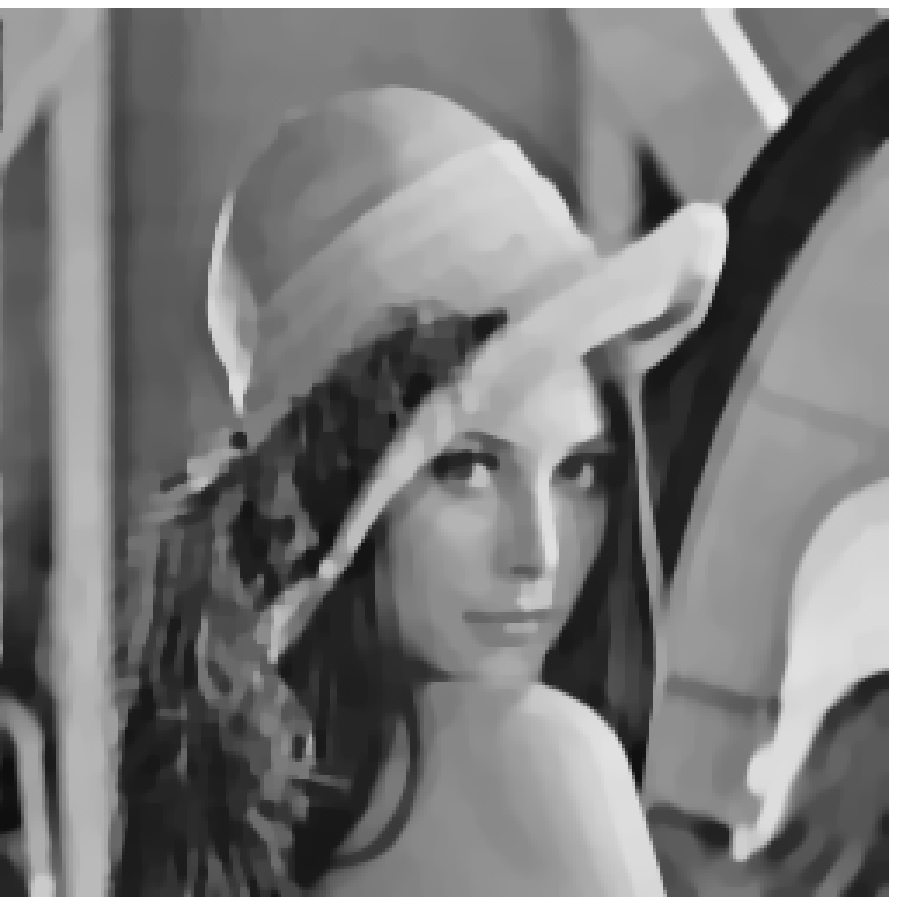} \\
(c) CTV, PSNR$=32.06$db      & (d) classical TV, PSNR$=28.33$db
\end{tabular}
}
\end{center}
\caption{{\protect\small (a) and (b) are the images restored from the test
image "Lena" convoluted with Gaussian blur kernel with $\protect\sigma_b=1$
by our method and by classical TV method respectively. (c) and (d) are the
images restored from the test image "Lena" convoluted with $9\times9$ box
average blur kernel by our method and by classical TV method respectively.}}
\label{Fig image}
\end{figure}

In the case of the Gaussian blur kernel, Figure \ref{Fig gau etv} (c) shows that
values of $||\mathcal{K}^{\ast }(\mathcal{K}f_{k}-u)||_{1}$ computed by the
new algorithm CTV and those by the classical TV algorithm are nearly the
same, when $\log k$ varies in the interval $[0,3]$. However, for $\log k$
varying in the interval $[3,8]$, the values of $||\mathcal{K}^{\ast }(%
\mathcal{K}f_{k}-u)||_{1}$ by the classical TV algorithm remain constant,
while those computed by our algorithm continue to decrease to $0$.
Accordingly, we see in Figure \ref{Fig gau etv} (d) that the PSNR (the
definition of PSNR see Section \ref{sec Computational algorithm}) values by
the classical TV algorithm and by the new proposed one, are almost the same
for $\log k$ varying in the interval $[0,3],$ imitating the evolution of $||%
\mathcal{K}^{\ast }(\mathcal{K}f_{k}-u)||_{1};$ however, for $\log k$
varying in the interval $[3,8]$, the PSNR value by the classical TV
algorithm does not increase any more, while the corresponding PSNR by the
new algorithm continues to rise. In the case of $9\times9$ box average blur
kernel, Figure \ref{Fig box etv} displays similar evolutions as those of the
Gaussian blur kernel case. Figure \ref{Fig image} shows the images restored
by our method and the classical TV method respectively. Our method keeps
much more details of the image than the classical TV method.

\subsection{Direct Gradient Descent}

\begin{figure}[tbp]
\begin{center}
\renewcommand{\arraystretch}{0.5} \addtolength{\tabcolsep}{-6pt} \vskip3mm {%
\fontsize{8pt}{\baselineskip}\selectfont
\begin{tabular}{cc}
\multicolumn{2}{c}{ CTV and DGD (Gaussian blur kernel)}\\
\includegraphics[width=0.49\linewidth]{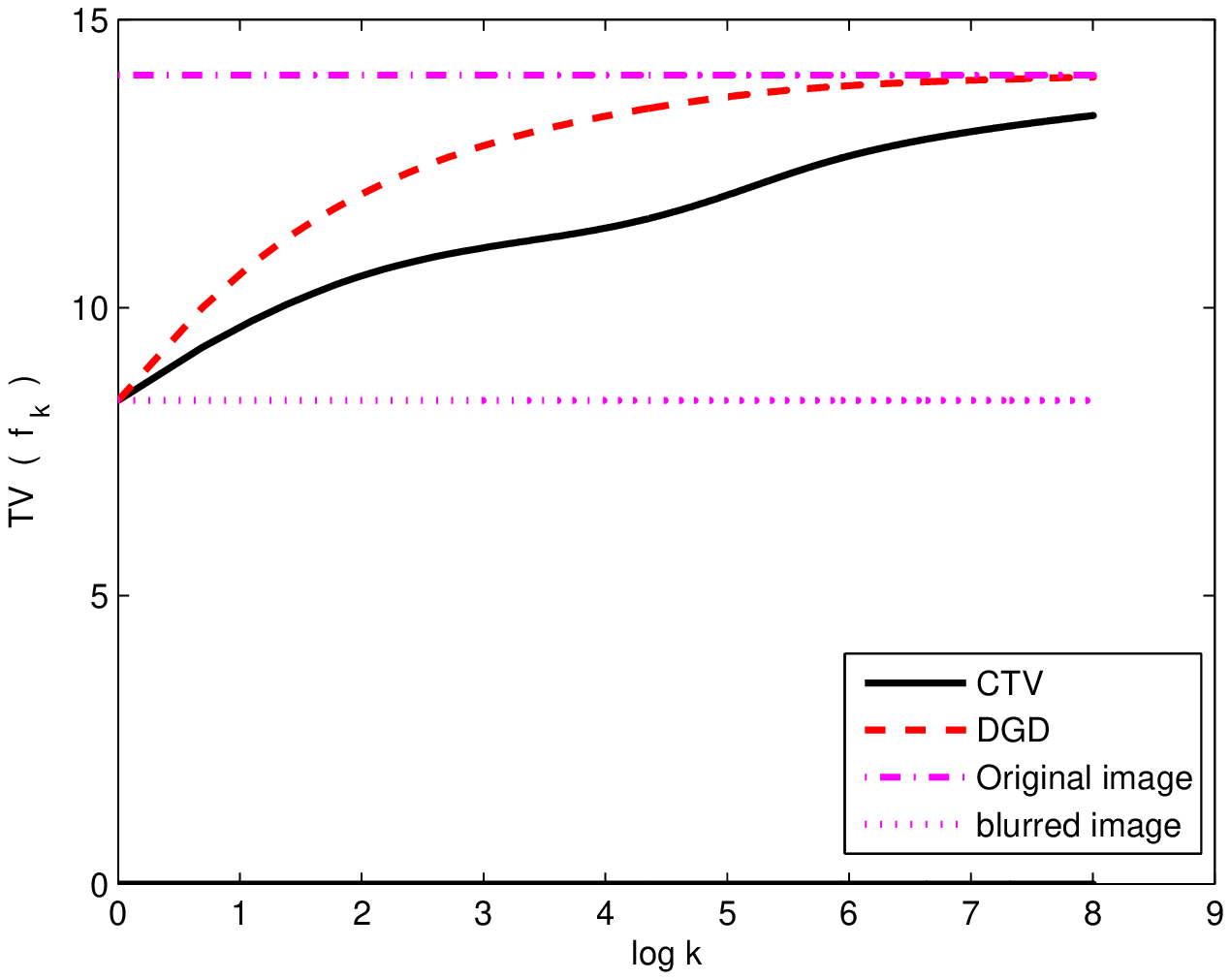} & %
\includegraphics[width=0.49\linewidth]{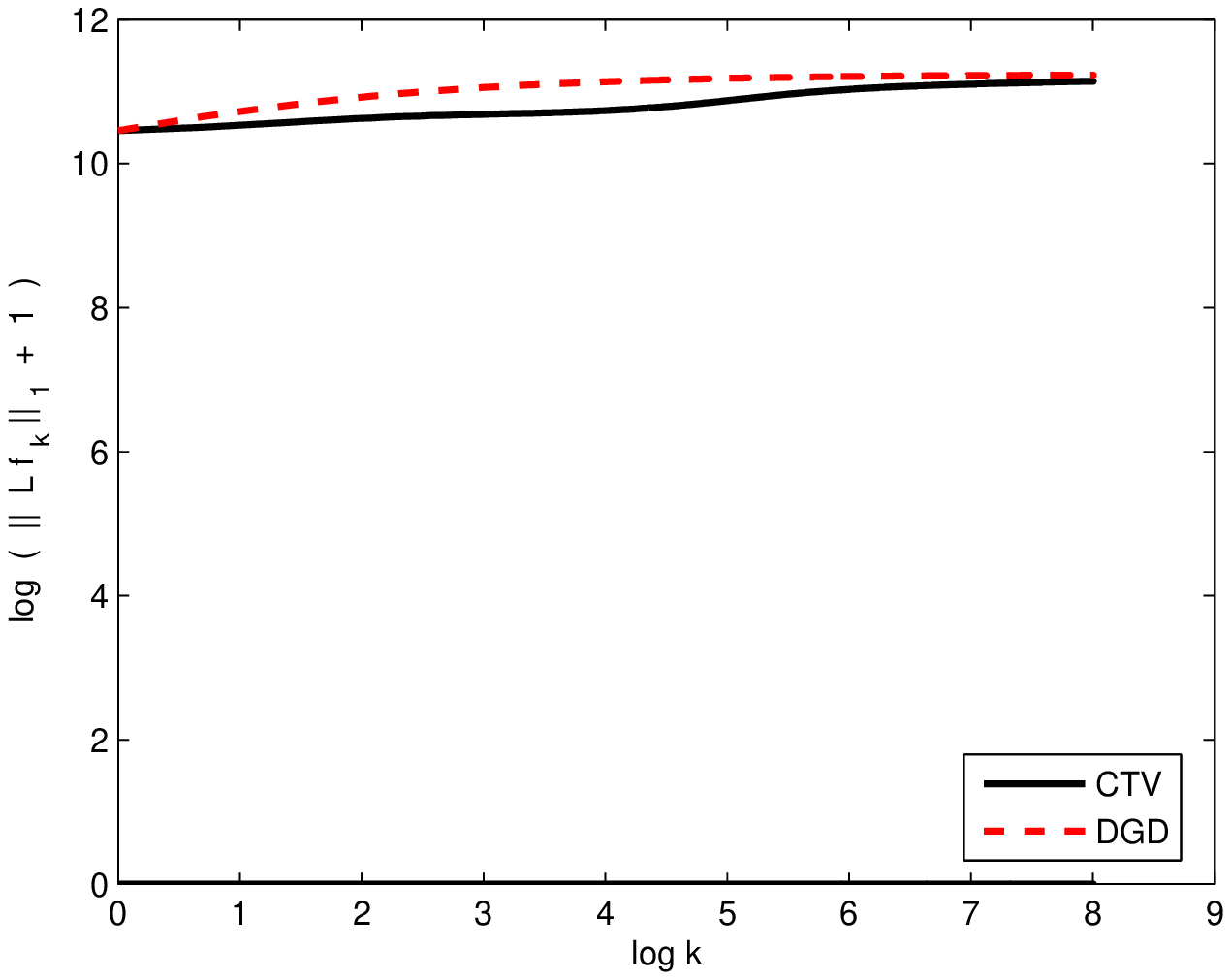} \\
(a) & (b) \\
\includegraphics[width=0.49\linewidth]{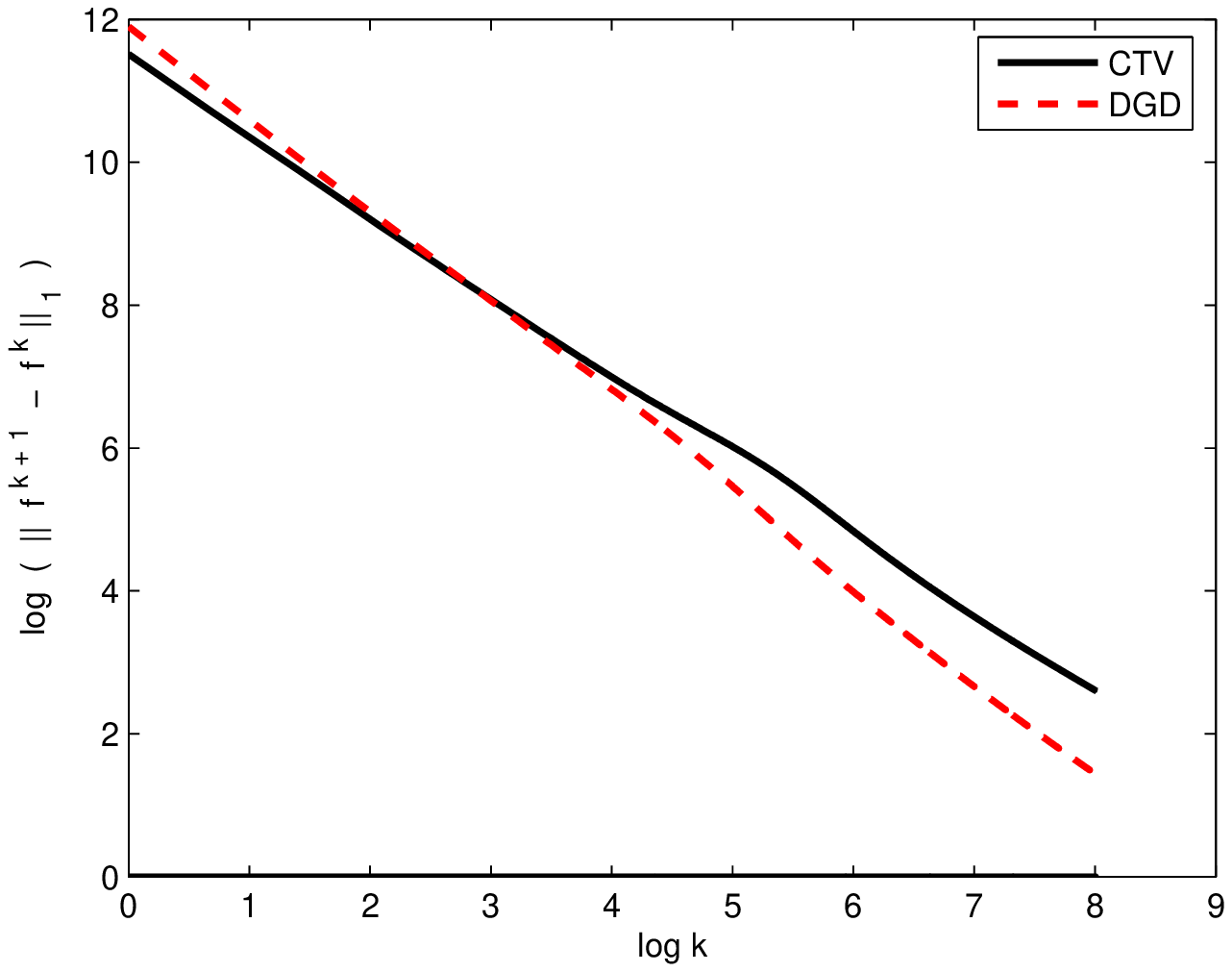} & %
\includegraphics[width=0.49\linewidth]{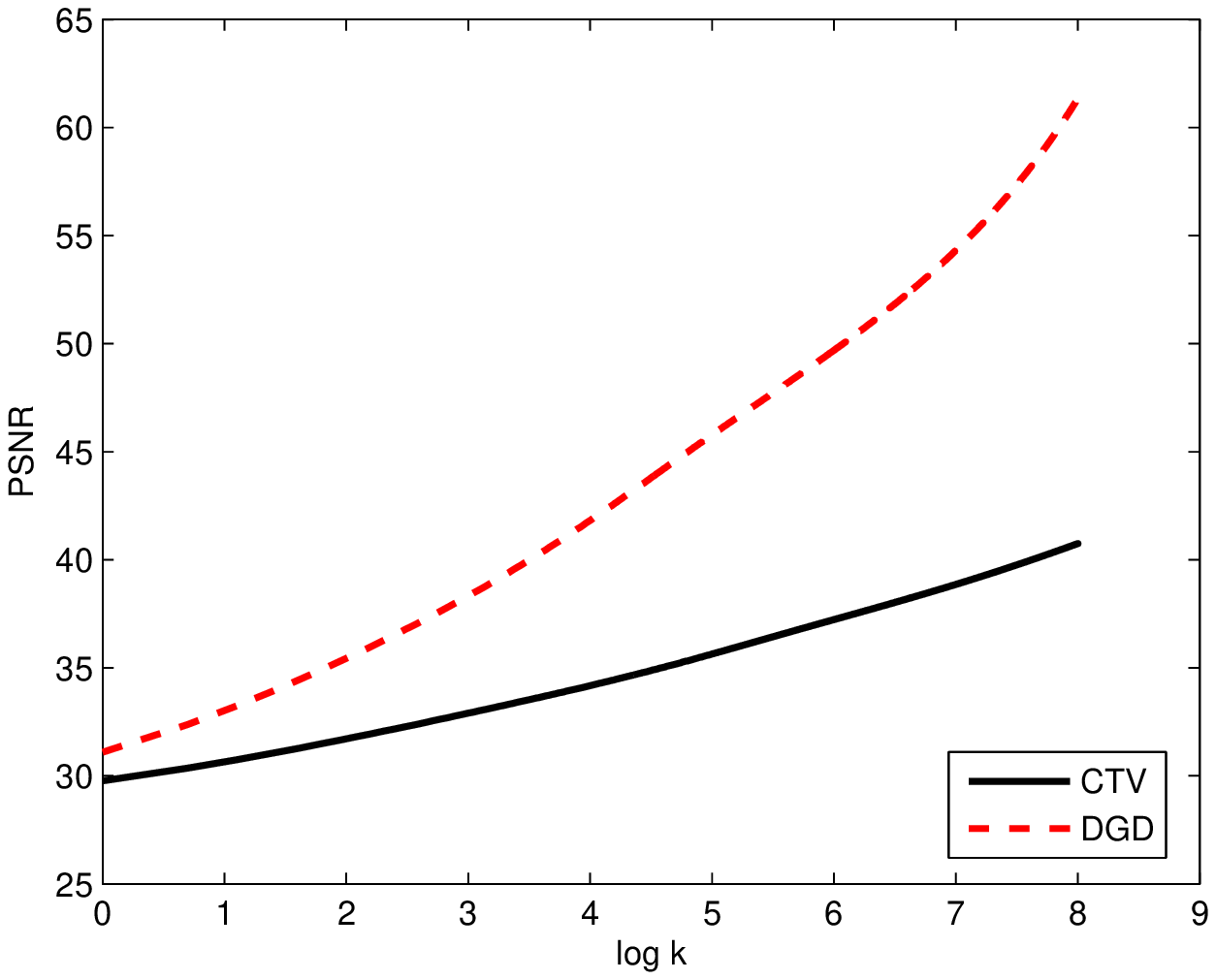} \\
(c) & (d)%
\end{tabular}
}
\end{center}
\caption{{\protect\small Comparison between CTV and DGD. The test image "Lena" is convoluted with Gaussian
blur kernel of standard deviation $\protect\sigma_b=1$. We choose $h =0.1$, $%
\protect\lambda=1$, $\max Iter=3000$ and $\protect\theta=0.98$. (a) The
evolution of $TV(f_k)$ as a function of $\log k$. (b) The evolution of $%
\log(||Lf_k||_1 + 1)$ as a function of $\log k$. (c) The evolution of $%
||f_{k + 1}-f_k||_1$ as a function of $\log k$, (d) The evolution of PSNR
value as a function of $\log k$. }}
\label{Fig gd gau}
\end{figure}

Suppose that $\mathcal{K}$ a linear operator from $R^{N^{2}}$ to $R^{N^{2}}.$
To solve the deconvolution problem (\ref{deconv approx}), inspired by (\ref%
{iteration jin}), we can also consider the following modified version, that
we call \textit{Direct Gradient Descent} (DGD):
\begin{equation}
f_{k + 1}=f_{k}-h\left( \theta _{k}Lf_{k} + \lambda (\mathcal{K}f_{k}-u)\right) ,
\label{iteration sp2}
\end{equation}%
where $0<\theta _{k}\rightarrow 0.$ If the algorithm converges, say $%
f_{k}\rightarrow \widetilde{f},$ then $\widetilde{f}$ satisfies exactly the
equation (\ref{deconv approx}). Therefore we could expect that this
algorithm gives a better solution to the deconvolution problem. By
simulations, this is shown to be the case when $\mathcal{K}$ is the Gaussian blur
kernel and the noise is absent. In this case we find that the DGD approach
restores the blurred image more efficiently than classical TV and the CTV
regularization approaches introduced above. The results of comparison
between the classical TV, CTV and DGD are displayed in Figures \ref{Fig gau
etv} and \ref{Fig gd gau}.

\begin{figure}[tbp]
\begin{center}
\renewcommand{\arraystretch}{0.5} \addtolength{\tabcolsep}{-6pt} \vskip3mm {%
\fontsize{8pt}{\baselineskip}\selectfont
\begin{tabular}{cc}
\multicolumn{2}{c}{ DGD with box average blur kenel}\\
\includegraphics[width=0.49\linewidth]{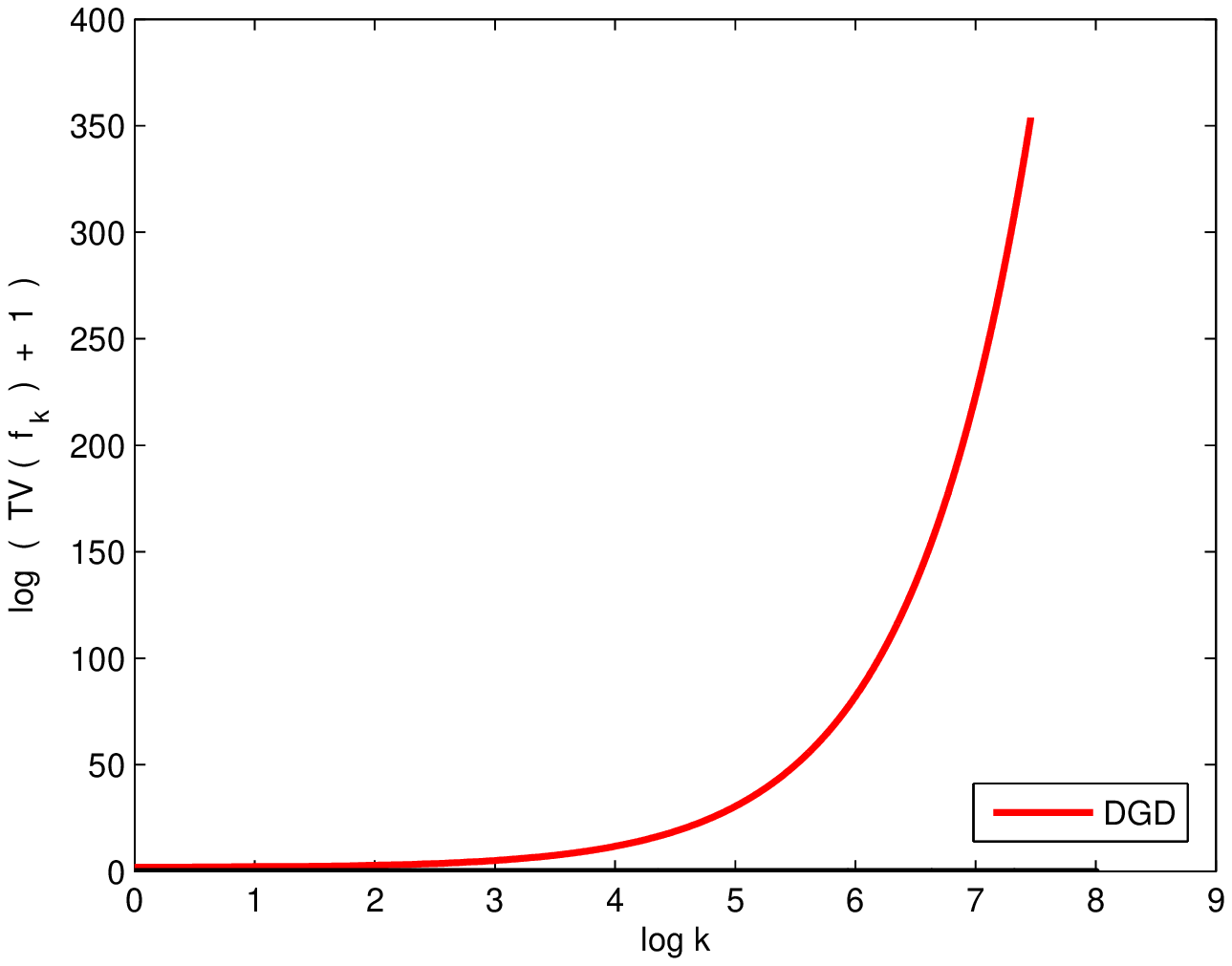} & %
\includegraphics[width=0.49\linewidth]{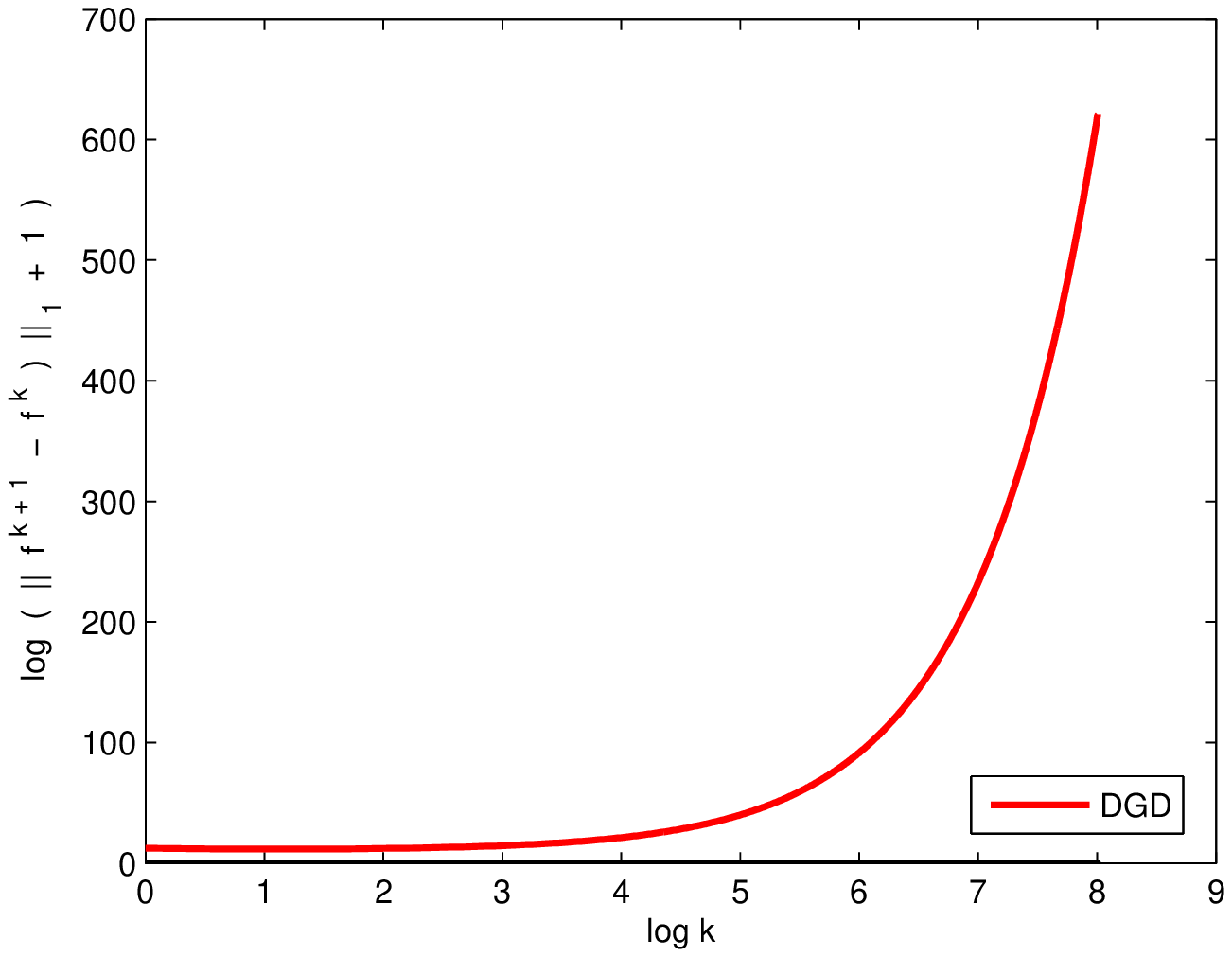} \\
(a) & (b)%
\end{tabular}
}
\end{center}
\caption{{\protect\small The test image "Lena" is convoluted with $9\times9$
box average blur kernel. We choose $h =0.1$, $\protect\lambda=1$, $\max
Iter=3000$ and $\protect\theta=0.998$. (a) The evolution of $\log (TV(f_k) + 1)$
as a function of $\log k$. (b) The evolution of $\log
(||f_{k + 1}-f_{k}||_{1} + 1)$ as a function of $\log k$. }}
\label{Fig gd box}
\end{figure}

However, our simulations also show that the algorithm DGD is very sensible
to the noise and often may not converge. For example, when $\mathcal{K}$ is the
Gaussian blur kernel, but in the presence of the noise, the method does not
give good restoration results; when $\mathcal{K}$ is the $9\times 9$ box average
kernel, the algorithm diverges and the restoration results are not good with
or without noise (see Figure \ref{Fig gd box}).

Notice also that the CTV regularization approach applies even if the linear
operator $\mathcal{K}$ is from $R^{N^{2}}$ to $R^{M}$, where $M$ is possibly different
from $N^{2},$ contrary to the DGD approach which applies only when $M=N^{2}.$

The aforementioned arguments show that the CTV regularization approach (\ref%
{iteration jin}) should be proffered to the Direct Gradient Descent
algorithm (\ref{iteration sp2}), especially when the blurring is accompanied
with a noise contamination.

\section{Numerical simulation}

\subsection{Computational algorithm\label{sec Computational algorithm}}

In this section we present some numerical results to compare different
deblurring methods. We shall use CTV regularization algorithm to restore
several standard blurred and noisy images and compare it to the Tikhonov  +
NL/H$^{1}$ (Tikhonov regularization followed by a Non-Local weighted $H^{1}$
approach) and Tikhonov  +  NL/TV (Tikhonov regularization followed by a
Non-Local weighted TV approach) algorithms presented in \citep{lou2010image}%
. When no noise is present, we shall also give simulation results by the DGD
approach.

\textbf{Algorithm :}\quad BM3D + CTV(OPF + CTV)

\textbf{Input:} Data $Y$ and the degradation model $\mathcal{K}$. \newline
\textbf{Step 1. Denoising. }

Remove Gaussian noise from $Y$ by BM3D \citep{dabov2007image} or by
Optimization Weight Filter \citep{JinGramaLiuowf} to  get the denoised image $u$.
\newline
\textbf{Step 2. Deconvolution.}

Set $\varepsilon =5e^{-5}$ and choose parameters $\lambda $, $h$, $\theta $,
$k$ and $maxIter=3000$.

\textbf{Initialization: }$k=0$ and $f_{0}=u.$

\textbf{(a)} \textbf{compute} $f_{k + 1}$ by the gradient descent equation (%
\ref{iteration sp2})

\textbf{(b)} \textbf{if} $\Vert \mathcal{K}^{\ast }(\mathcal{K}f_{k + 1}-u)-%
\mathcal{K}^{\ast }(\mathcal{K}f_{k}-u)\Vert _{2}>\varepsilon $ and $%
k<maxIter$ \textbf{set} $k=k + 1$ \textbf{and go to (a).}

\textbf{Output:} final solution $f_{k + 1}$.

Note that  If there is no noise, we just do Step 2.

In the algorithm we use the control condition $\Vert \mathcal{K}^{\ast }(%
\mathcal{K}f_{k + 1}-u)-\mathcal{K}^{\ast }(\mathcal{K}f_{k}-u)\Vert _{2}\leq
\varepsilon $ since we would like to ensure that $\mathcal{K}^{\ast }(%
\mathcal{K}f_{k}-u)\rightarrow 0.$ We could replace this condition by $\Vert
f_{k + 1}-f_{k}\Vert _{2}\leq \varepsilon $.

The performance of the CTV regularization algorithm is measured by the usual
Peak Signal-to-Noise Ratio (PSNR) in decibels (db) defined as%
\begin{equation*}
PSNR=10\log _{10}\frac{255^{2}}{MSE},
\end{equation*}%
where
\begin{equation*}
MSE=\frac{1}{\mathrm{card}\,\mathbf{I}}\sum\limits_{x\in \mathbf{I}}(f(x)-%
\widetilde{f}(x))^{2}
\end{equation*}%
with $f$  the original image and $\widetilde{f}$ the estimated one.

\begin{table*}[tbp]
\caption{ The statistics of blur and noise we add to the images}
\label{Tab test image}
\begin{center}
\renewcommand{\arraystretch}{0.6} \vskip3mm {\fontsize{8pt}{\baselineskip}%
\selectfont
\begin{tabular}{lllll}
\hline
Image & Blur kernel & Gaussian noise 1 & Gaussian noise 2 & Gaussian noise 3
\\ \hline
Shape & Gaussian with $\sigma_b=2$ & $\sigma_n=0$ & $\sigma_n=10$ & $%
\sigma_n=20$ \\
Lena & Gaussian with $\sigma_b=1$ & $\sigma_n=0$ & $\sigma_n=10$ & $%
\sigma_n=20$ \\
Barbara & Gaussian with $\sigma_b=1$ & $\sigma_n=0$ & $\sigma_n=5$ & $%
\sigma_n=20$ \\
Carmeraman & $9\times9$ box average & $\sigma_n=0$ & $\sigma_n=3$ & $%
\sigma_n=10$ \\ \hline
\end{tabular}
} \vskip1mm
\end{center}
\end{table*}

\begin{figure}[tbp]
\begin{center}
\renewcommand{\arraystretch}{0.5} \addtolength{\tabcolsep}{-5pt} \vskip3mm {%
\fontsize{8pt}{\baselineskip}\selectfont
\begin{tabular}{cccc}
\includegraphics[width=0.23\linewidth]{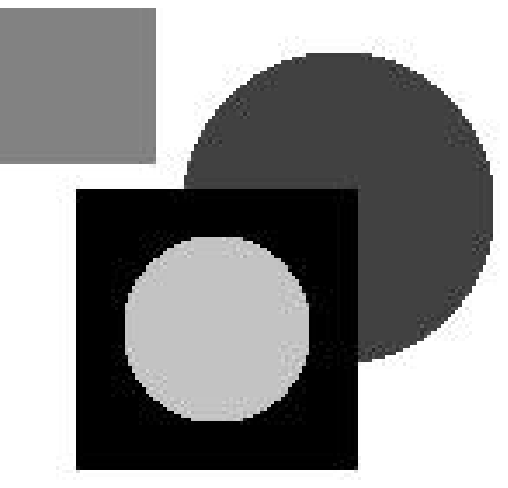} & %
\includegraphics[width=0.23\linewidth]{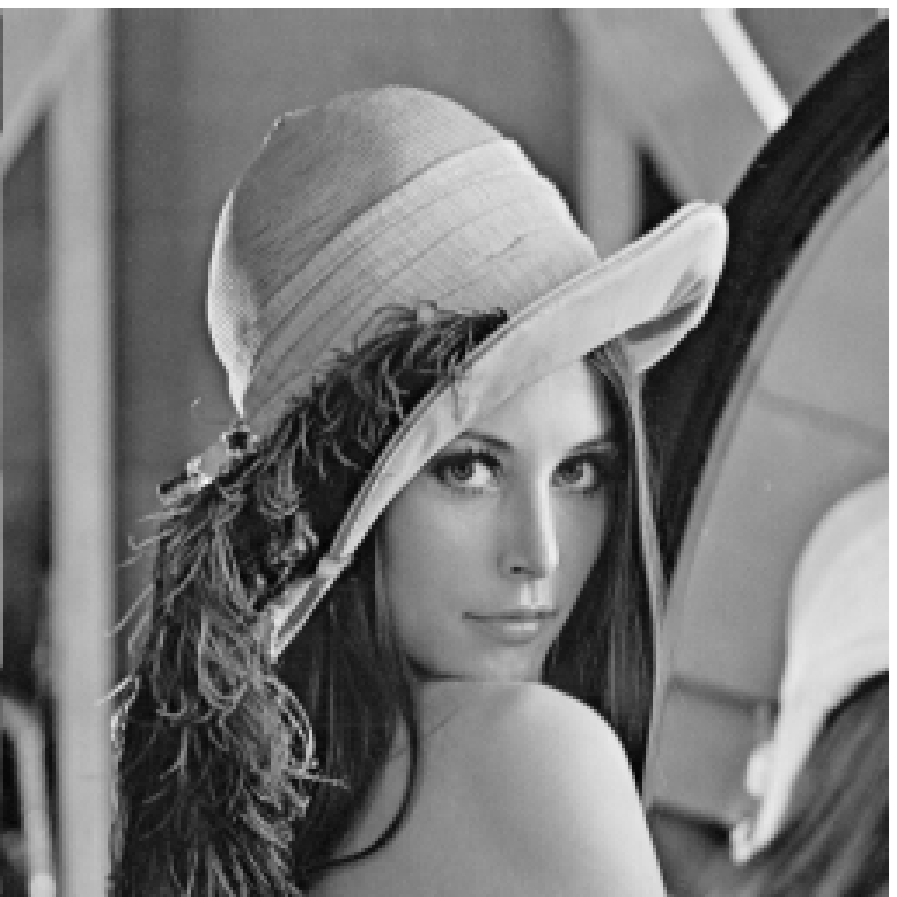} & %
\includegraphics[width=0.23\linewidth]{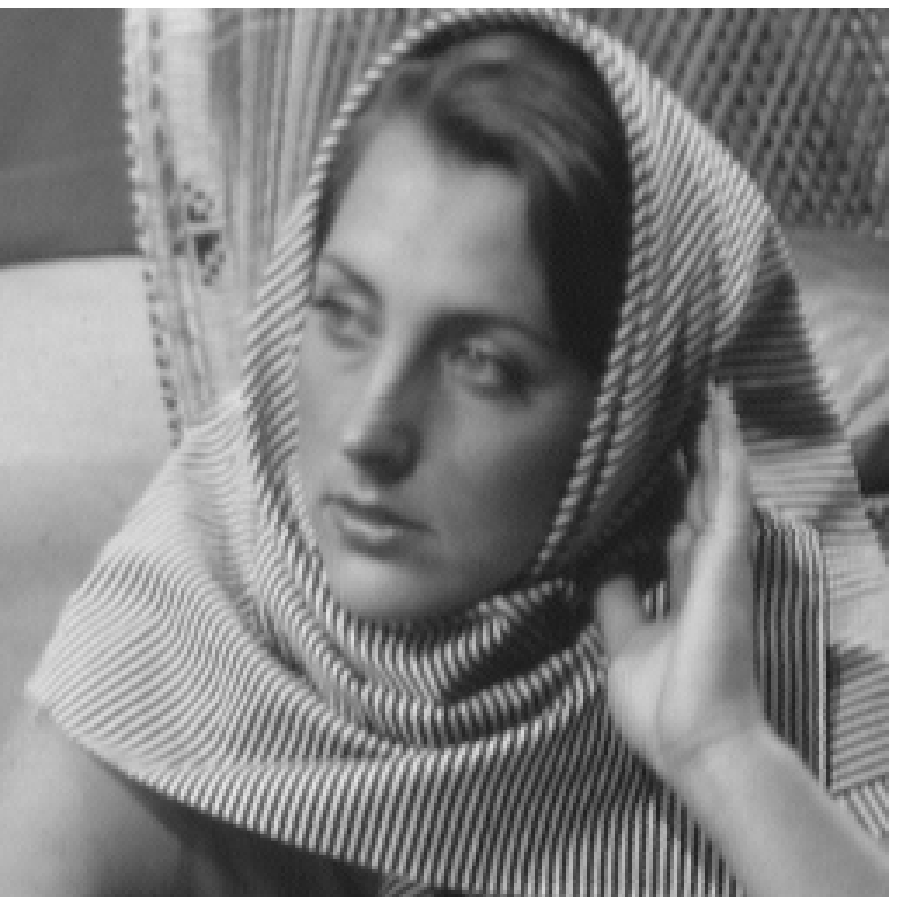} & %
\includegraphics[width=0.23\linewidth]{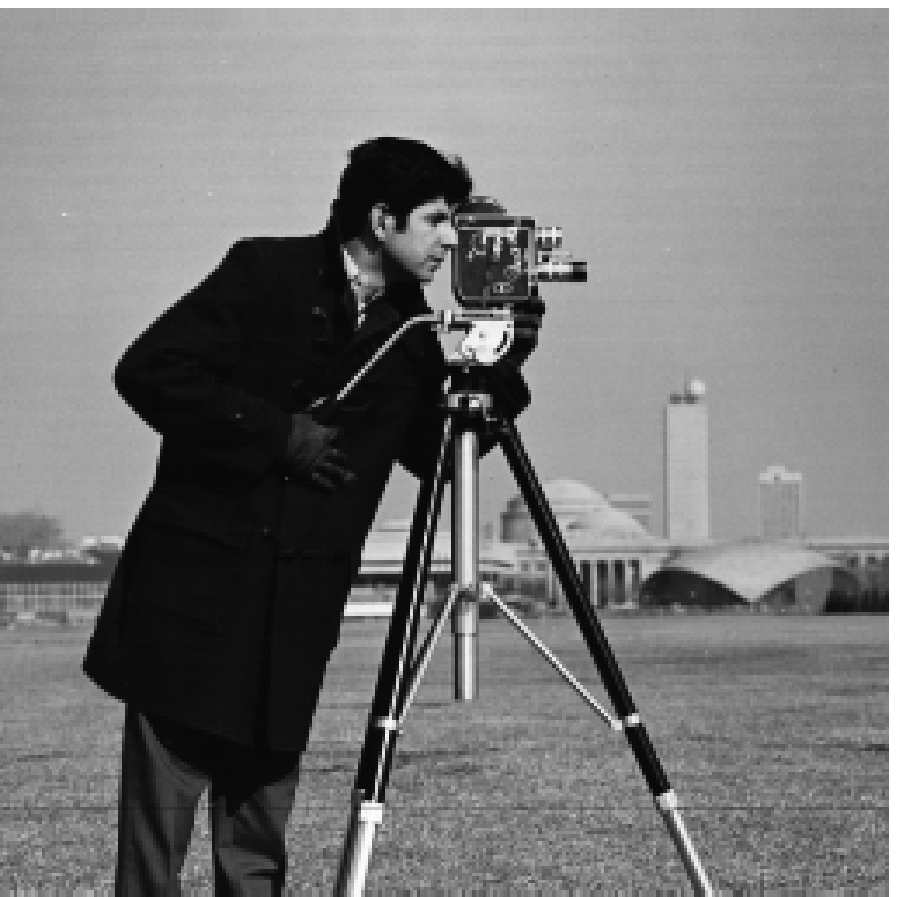} \\
(a) Shape $150\times150$ & (b)Lena $256\times256$ & (c) Barbara $%
256\times256 $ & (d) Cameraman $256\times256$%
\end{tabular}
}
\end{center}
\caption{{\protect\small The test images}}
\label{Fig test image}
\end{figure}

We test all the methods on four images: a synthetic image, Lena, Barbara and
Cameraman (see Figure \ref{Fig test image}) with several kinds of blur
kernels and several values of noise variances as listed in Table \ref{Tab
test image}. The synthetic image is referred to as Shape since it contains
geometric figures. The Cameraman image is of high contrast and has many
edges, while the Lena and Barbara images contain more textures.

\subsection{Simulation results}

We use BM3D or Optimization Weights Filter to remove Gaussian noise. Then we
use our CTV algorithm to inverse the denoised blurred image. For each
method, we present the PSNR results along with the applied parameters. We
list the PSNR values of the used methods in Table \ref{Tab simulation
results}. The Direct Gradient Descent method inverse excellently the
convoluted image with Gaussian blur kernel without any noise, and the PSNR
values are much higher than PSNR values of the other algorithms. However,
the method works well only for the Gaussian blur kernel without any noise.
For the noisy and blurred images, the method BM3D + CTV is the best way to
reconstruct these images. Figures \ref{Fig shape}, \ref{Fig lena}, \ref{Fig
Barbara} and \ref{Fig carmeraman} display the reconstructed images restored
by the different methods and the square of the difference between the
restored and original images. Figures \ref{Fig shape}, \ref{Fig lena} and %
\ref{Fig carmeraman} show that the method BM3D + CTV is the best one to remove
the noise and blur. From Figure \ref{Fig Barbara} we see that BM3D + CTV works
very well in reconstructing the image blurred with Gaussian blur kernel
without any noise, but the method DGD can work much better than BM3D + CTV.

\begin{table*}[tbp]
\caption{ The PSNR values of the simulation results for different methods}
\label{Tab simulation results}
\begin{center}
\renewcommand{\arraystretch}{0.6} \vskip3mm {\fontsize{8pt}{\baselineskip}%
\selectfont
\begin{tabular}{lllrrrrrr}
\hline
Image & Blur kernel & Noise & Tikhonov  & Tikhonov  & OPF & BM3D &
DGD &    \\
      &             &       &   + NL/H$^1$ &   + NL/TV &  + CTV &  + CTV &
    &    \\\hline
&  & $\sigma_n=0$ & 23.62db & 23.62db & 32.28db & 32.28db & 35.49db &   \\
Shape & Gaussian with $\sigma_b=2$ & $\sigma_n=10$ & 27.53db & 28.71db &
29.13db & 29.15db & ---.---db &    \\
&  & $\sigma_n=20$ & 25.02db & 26.58db & 27.75db & 27.50db & ---.---db &
\\ \hline
&  & $\sigma_n=0$ & 37.80db & 37.80db & 42.05db & 42.05db & 53.53db &    \\
Lena & Gaussian with $\sigma_b=1$ & $\sigma_n=10$ & 27.56db & 27.55db &
29.08db & 29.53db & ---.---db &    \\
&  & $\sigma_n=20$ & 25.29db & 25.26db & 27.48db & 27.84db & ---.---db &
\\ \hline
&  & $\sigma_n=0$ & 40.86db & 40.86db & 45.45db & 45.45db & 56.82db &    \\
Barbara & Gaussian with $\sigma_b=1$ & $\sigma_n=5$ & 28.17db & 28.25db &
29.48db & 30.35db & ---.---db &    \\
&  & $\sigma_n=20$ & 23.31db & 23.36db & 24.67db & 25.45db & ---.---db &
\\ \hline
&  & $\sigma_n=0$ & 27.22db & 27.22db & 30.37db & 30.37db & ---.---db &
\\
Carmeraman & $9\times9$ box average & $\sigma_n=3$ & 25.40db & 25.43db &
25.78db & 26.07db & ---.---db &    \\
&  & $\sigma_n=20$ & 22.10db & 22.28db & 22.58db & 22.80db & ---.---db &
\\ \hline
\end{tabular}
} \vskip1mm
\end{center}
\end{table*}

\begin{figure*}[tbp]
\begin{center}
\renewcommand{\arraystretch}{0.5} \addtolength{\tabcolsep}{-2pt} \vskip3mm {%
\fontsize{8pt}{\baselineskip}\selectfont
\begin{tabular}{cc}
Original and degraded image & Tikhonov  + NL/H$^1,$ PSNR$=27.53$db\\
\includegraphics[width=0.22\linewidth]{shape150.eps} %
\includegraphics[width=0.22\linewidth]{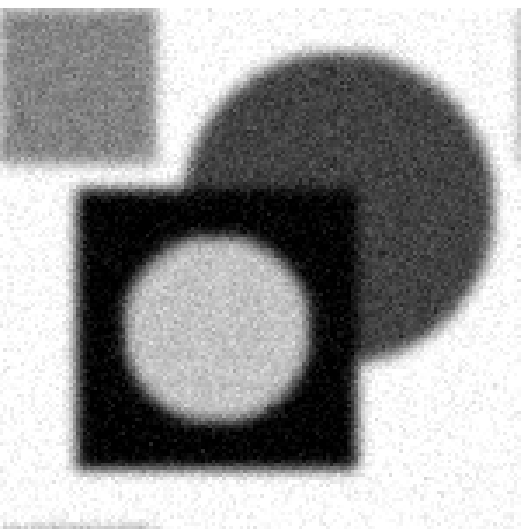} & %
\includegraphics[width=0.22\linewidth]{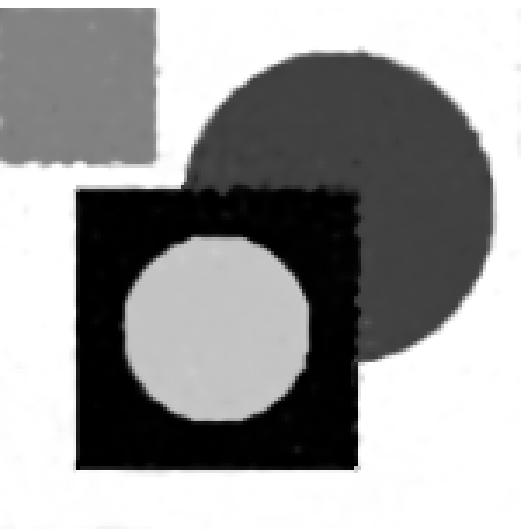} %
\includegraphics[width=0.22\linewidth]{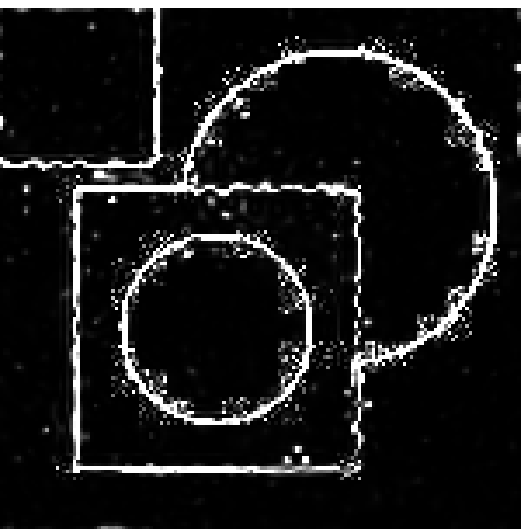} \\
 Tikhonov  + NL/TV, PSNR$=28.71$db & OPF + CTV, PSNR$=29.13$db \\
\includegraphics[width=0.22\linewidth]{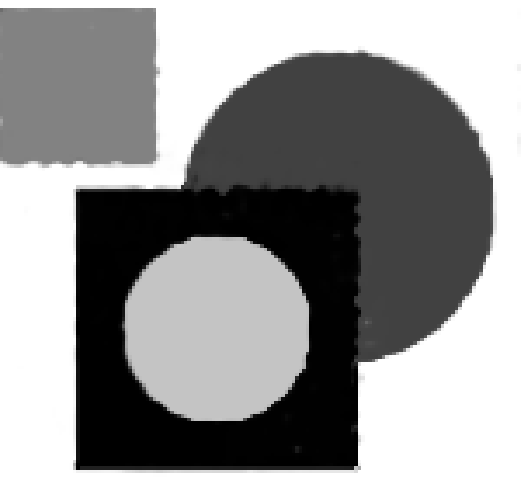} %
\includegraphics[width=0.22\linewidth]{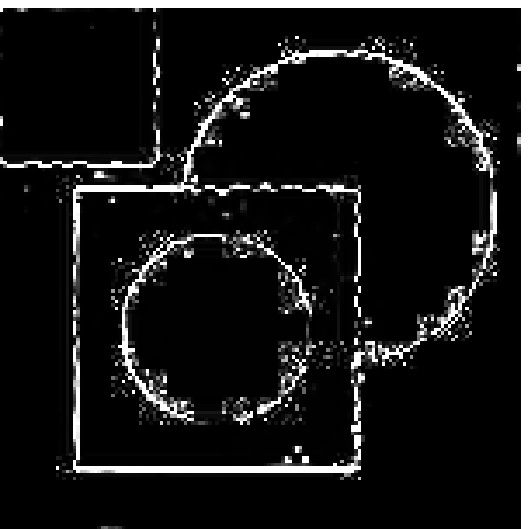} & %
\includegraphics[width=0.22\linewidth]{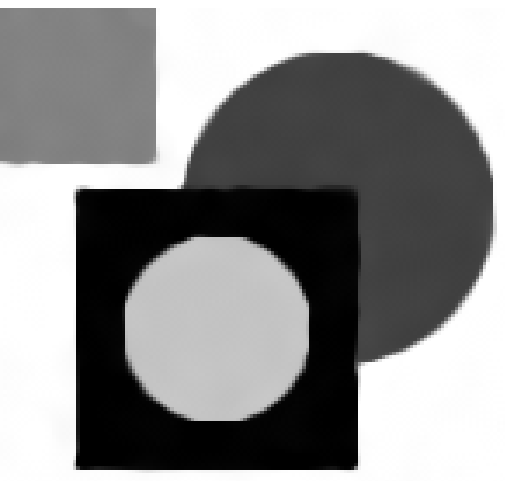} %
\includegraphics[width=0.22\linewidth]{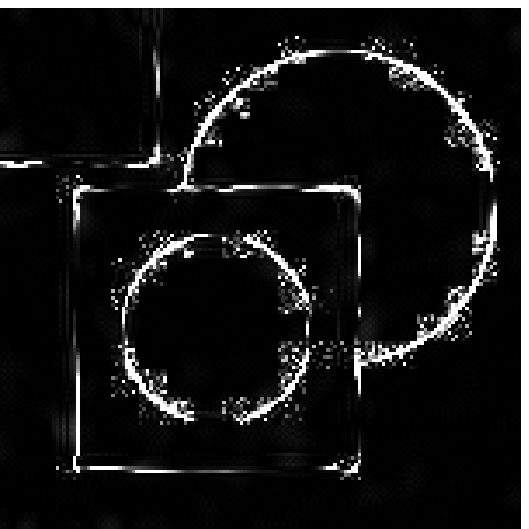} \\
&  \\
BM3D + CTV, PSNR$=29.15$db &  \\
\includegraphics[width=0.22\linewidth]{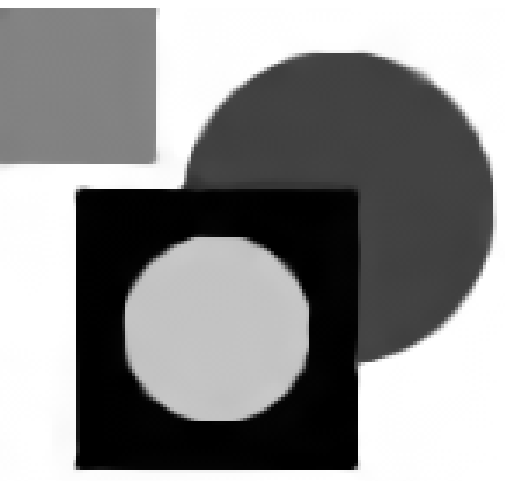} %
\includegraphics[width=0.22\linewidth]{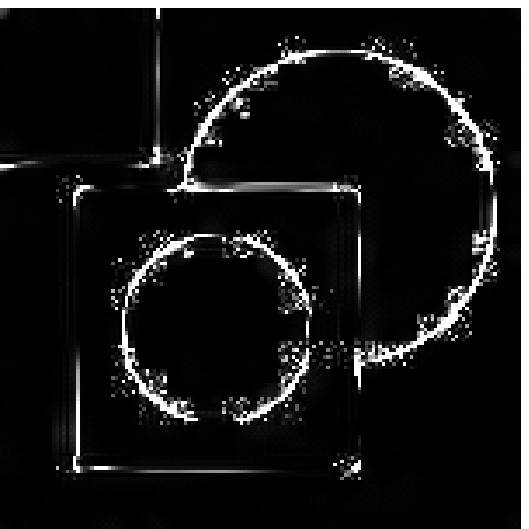} &
\end{tabular}
} \vskip1mm
\par
\rule{0pt}{-0.2pt}%
\par
\vskip1mm 
\end{center}
\caption{{\protect\small A $150\times150$ image with Gaussian blur $\protect%
\sigma_b=2$ and Gaussian noise $\protect\sigma_n=10$, the reconstructed
image by different methods and their square errors.}}
\label{Fig shape}
\end{figure*}

\begin{figure*}[tbp]
\begin{center}
\renewcommand{\arraystretch}{0.5} \addtolength{\tabcolsep}{-2pt} \vskip3mm {%
\fontsize{8pt}{\baselineskip}\selectfont
\begin{tabular}{cc}
Original and degraded image & Tikhonov  + NL/H$^1$, PSNR$=27.56$db \\
\includegraphics[width=0.22\linewidth]{lena.eps} \includegraphics[width=0.22%
\linewidth]{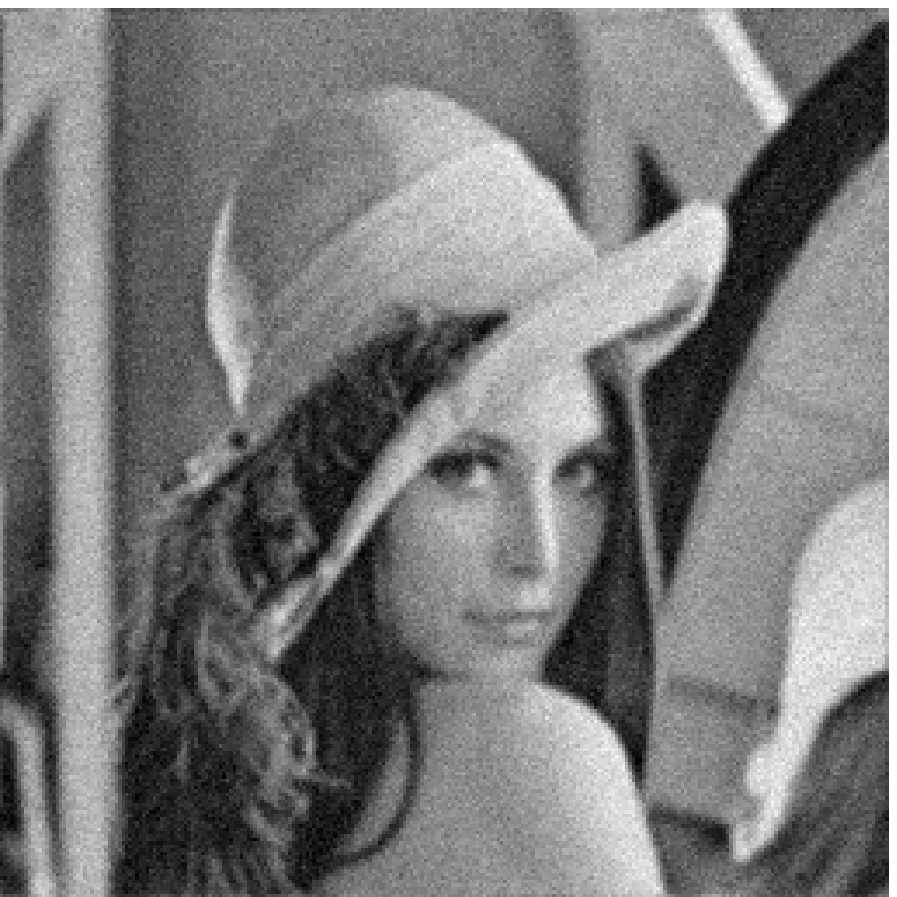} & \includegraphics[width=0.22%
\linewidth]{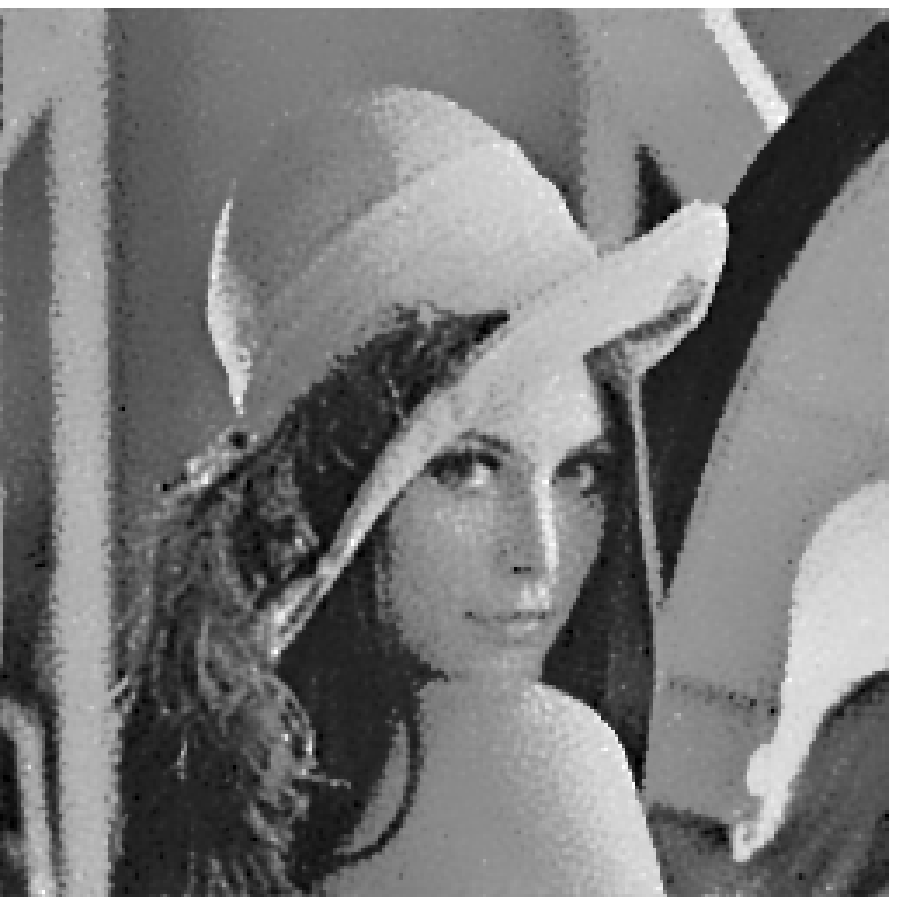} \includegraphics[width=0.22%
\linewidth]{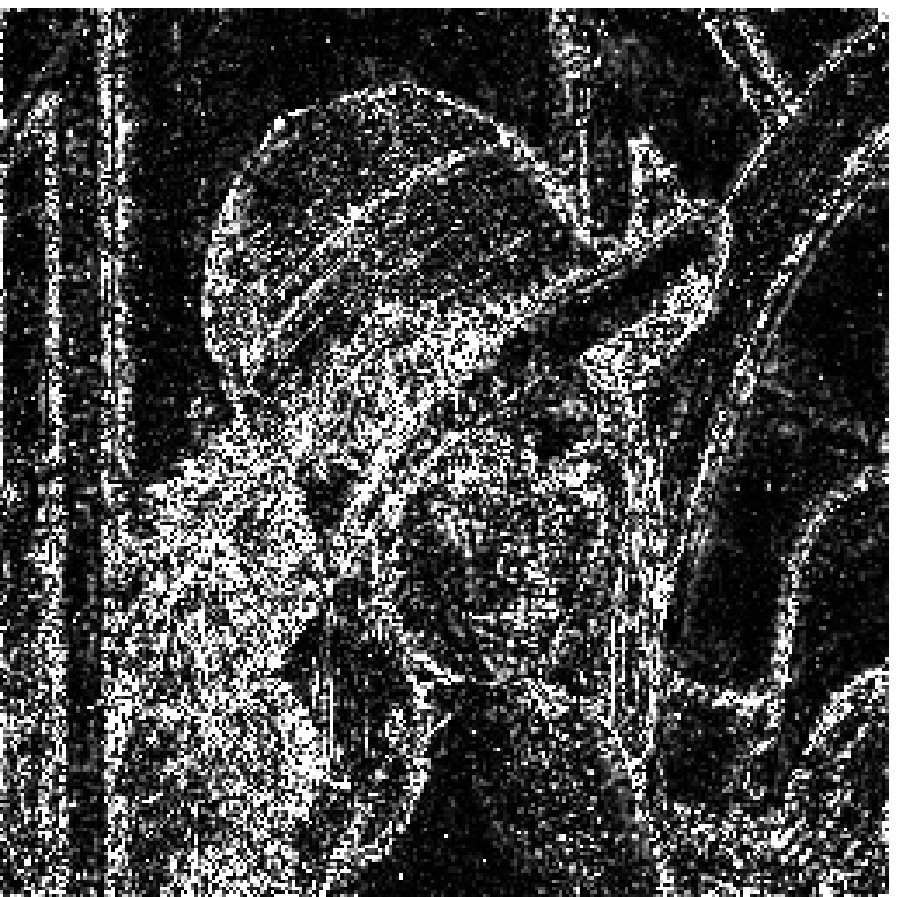} \\
 Tikhonov  + NL/TV, PSNR$=27.55$db & OPF + CTV, PSNR$=29.08$db \\
\includegraphics[width=0.22\linewidth]{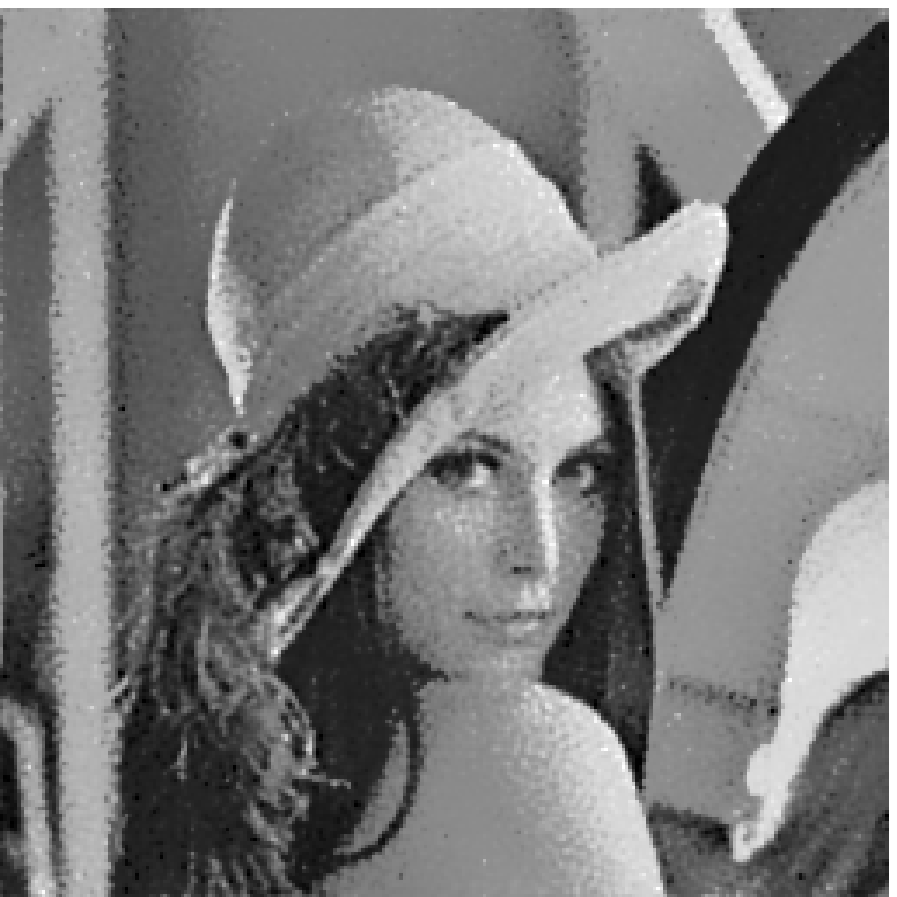} %
\includegraphics[width=0.22\linewidth]{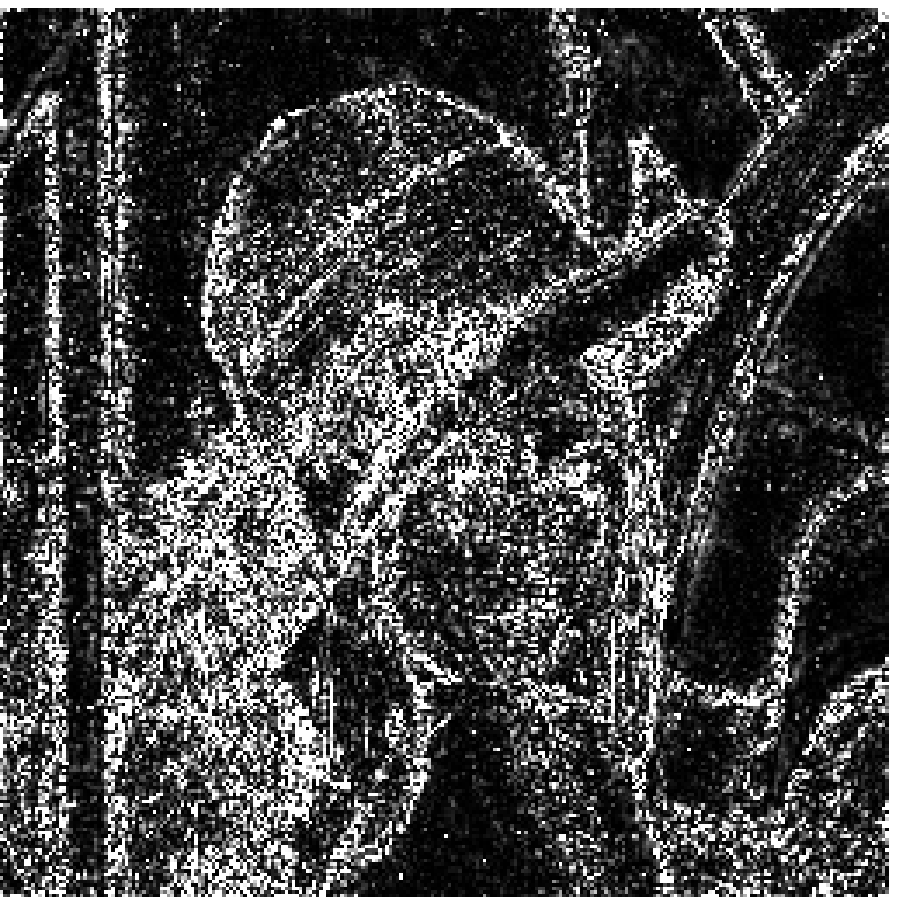} & %
\includegraphics[width=0.22\linewidth]{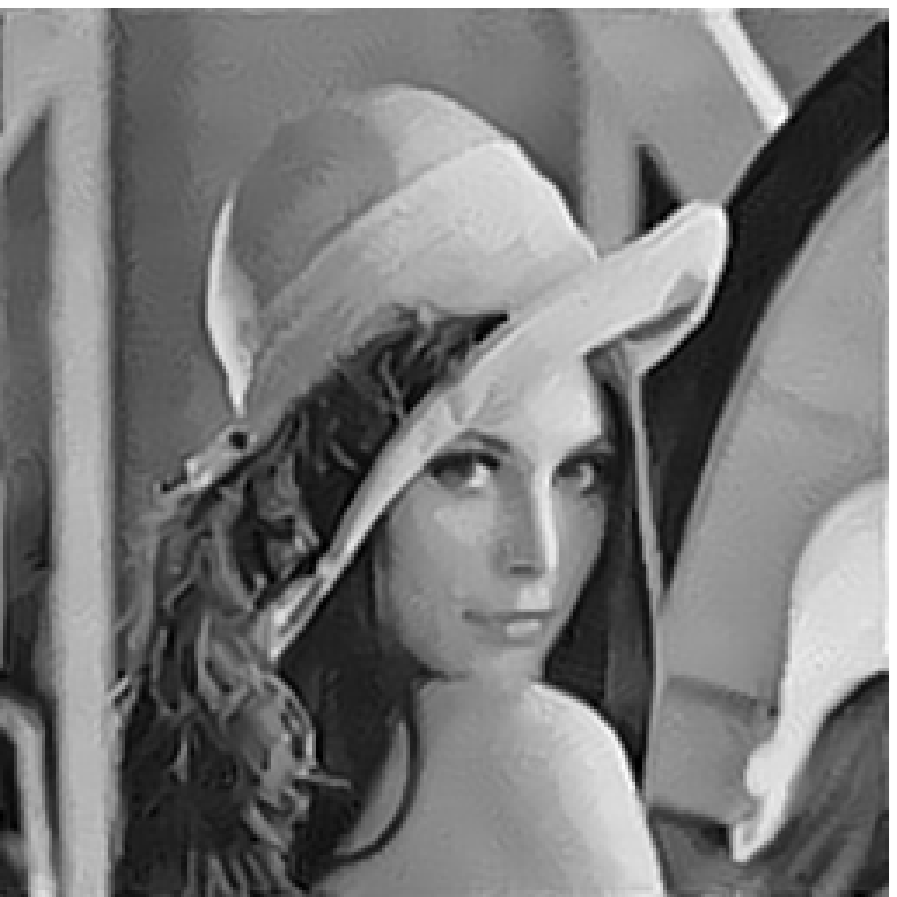} %
\includegraphics[width=0.22\linewidth]{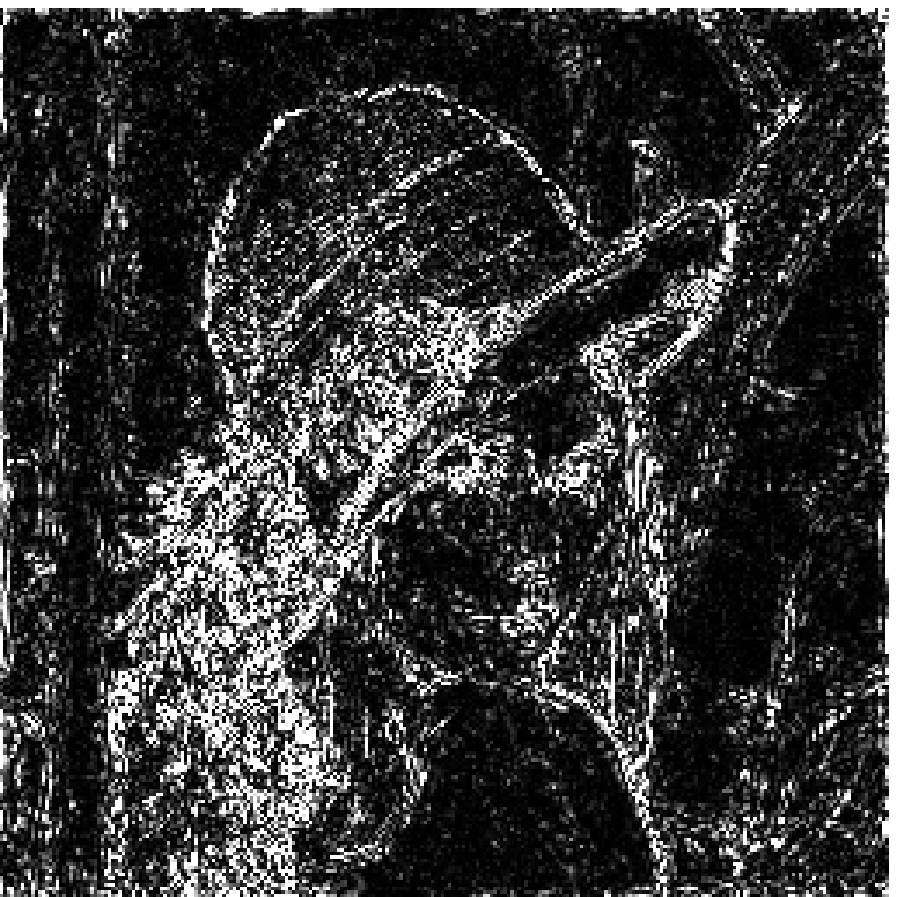} \\
BM3D + CTV, PSNR$=29.53$db &  \\
\includegraphics[width=0.22\linewidth]{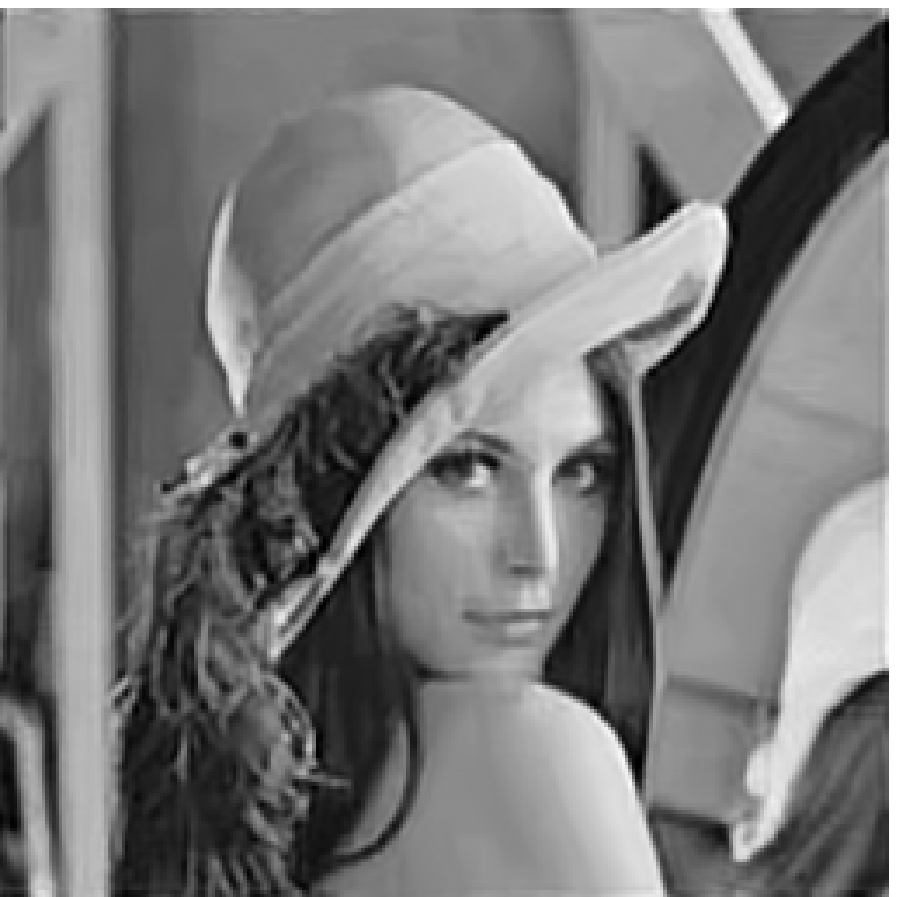} %
\includegraphics[width=0.22\linewidth]{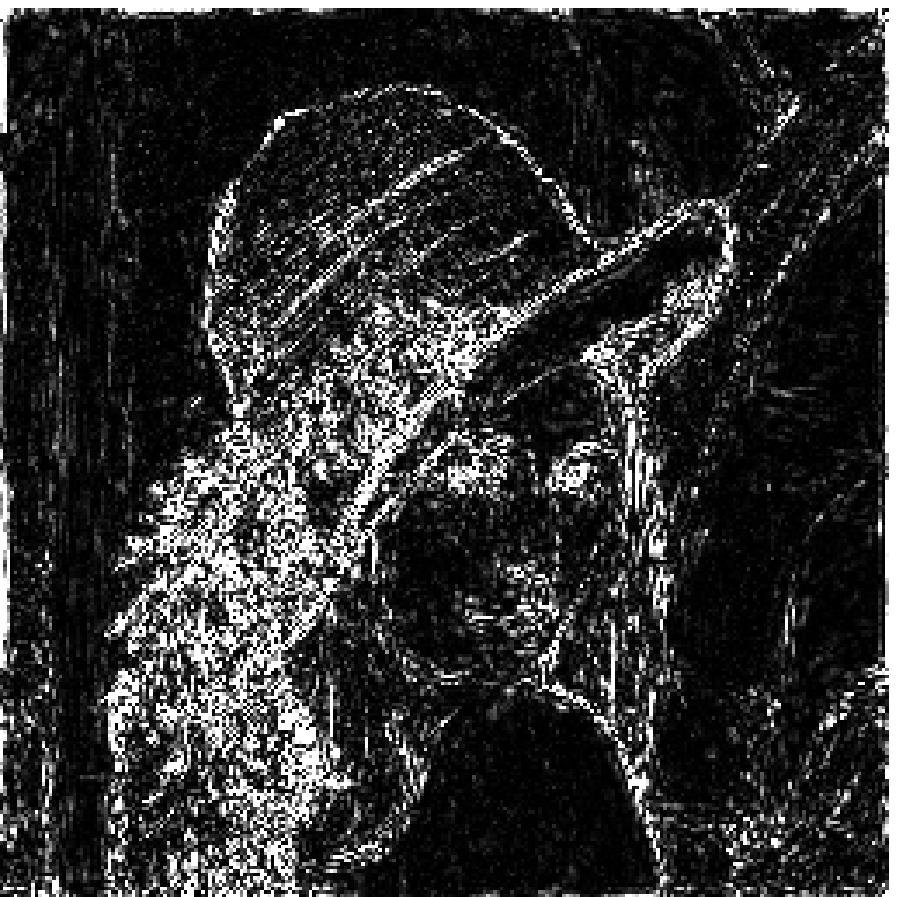} &
\end{tabular}
} \vskip1mm
\par
\rule{0pt}{-0.2pt}%
\par
\vskip1mm 
\end{center}
\caption{{\protect\small Lena $256\times256$ with Gaussian blur $\protect%
\sigma_b=1$ and Gaussian noise $\protect\sigma_n=10$, the reconstructed
image by different methods and their square errors.}}
\label{Fig lena}
\end{figure*}

\begin{figure*}[tbp]
\begin{center}
\renewcommand{\arraystretch}{0.5} \addtolength{\tabcolsep}{-2pt} \vskip3mm {%
\fontsize{8pt}{\baselineskip}\selectfont
\begin{tabular}{cc}
Original and degraded image & Tikhonov  + NL/H$^1$, PSNR$=40.86$db \\
\includegraphics[width=0.22\linewidth]{barbara256.eps} %
\includegraphics[width=0.22\linewidth]{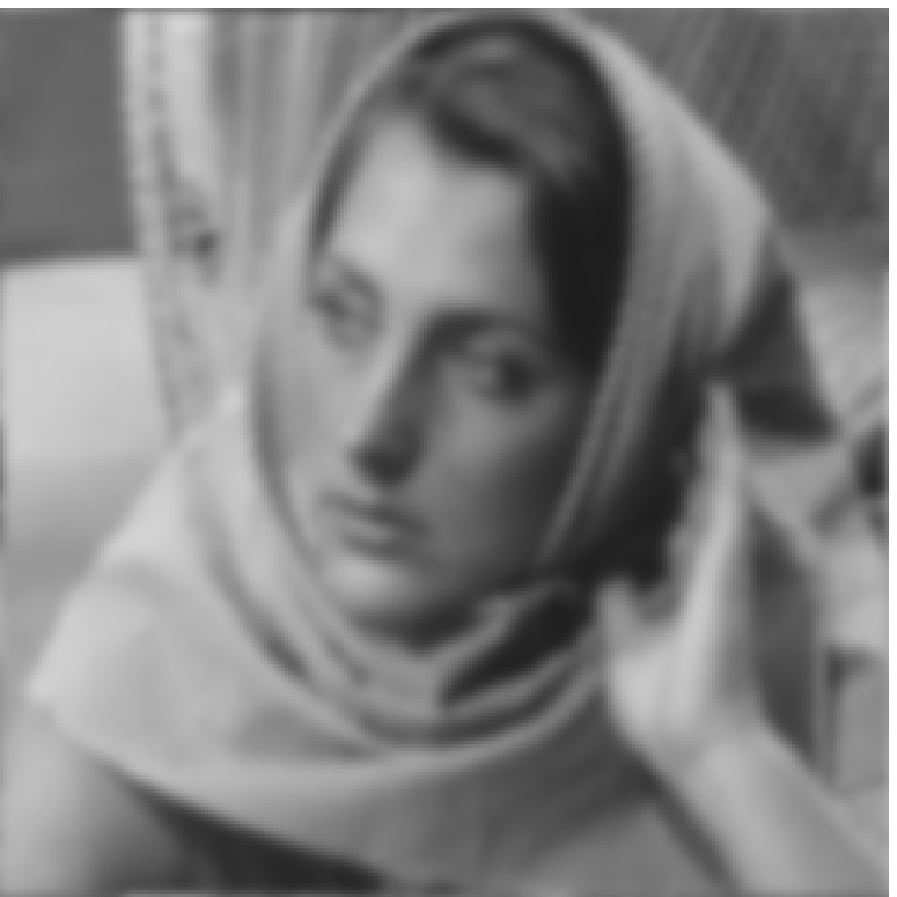} & %
\includegraphics[width=0.22\linewidth]{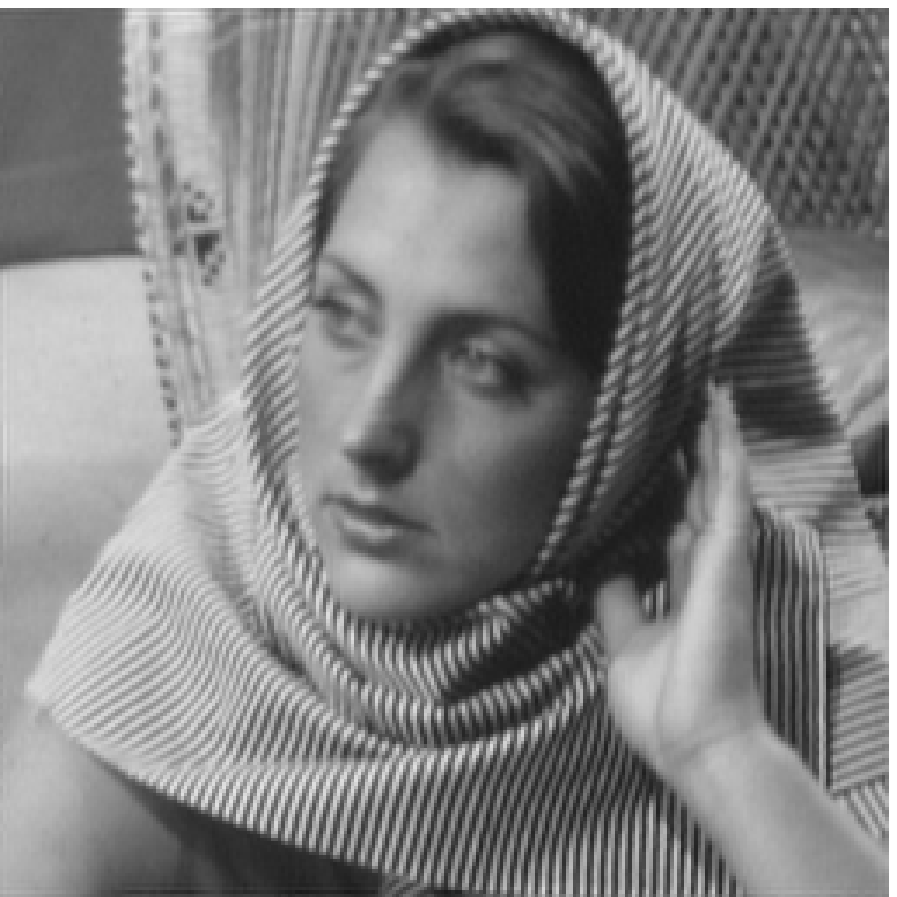} %
\includegraphics[width=0.22\linewidth]{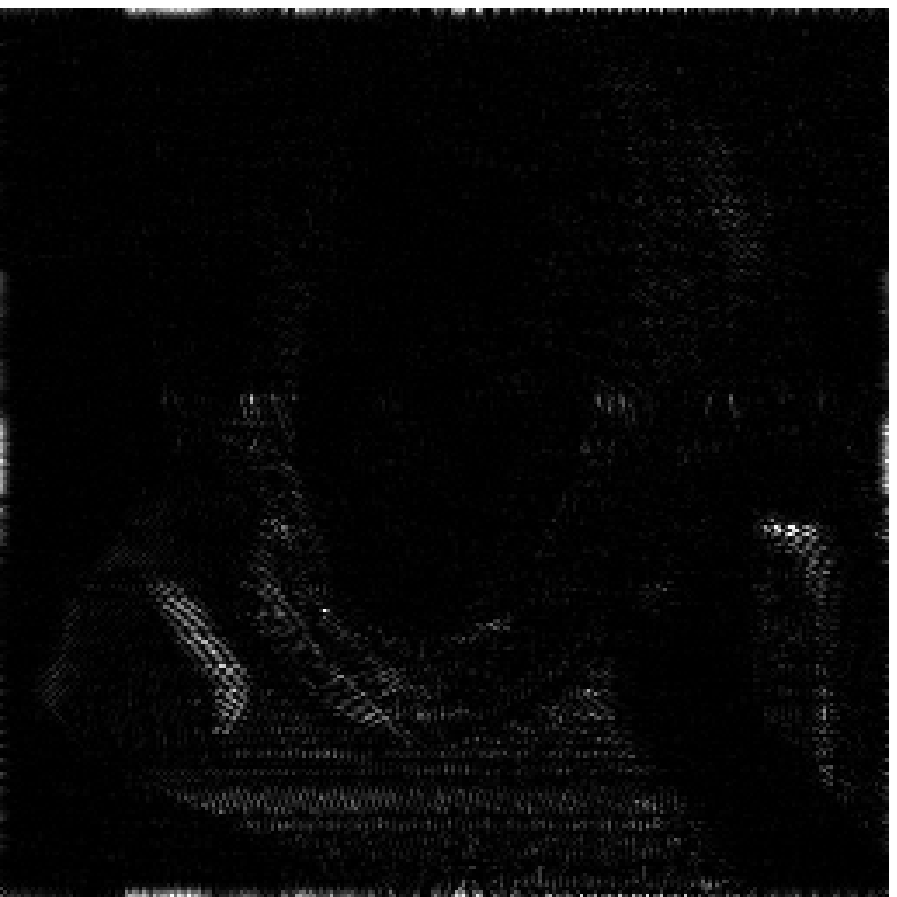} \\
 Tikhonov  + NL/TV, PSNR$=40.86$db & OPF + CTV, PSNR$=45.45$db \\
\includegraphics[width=0.22\linewidth]{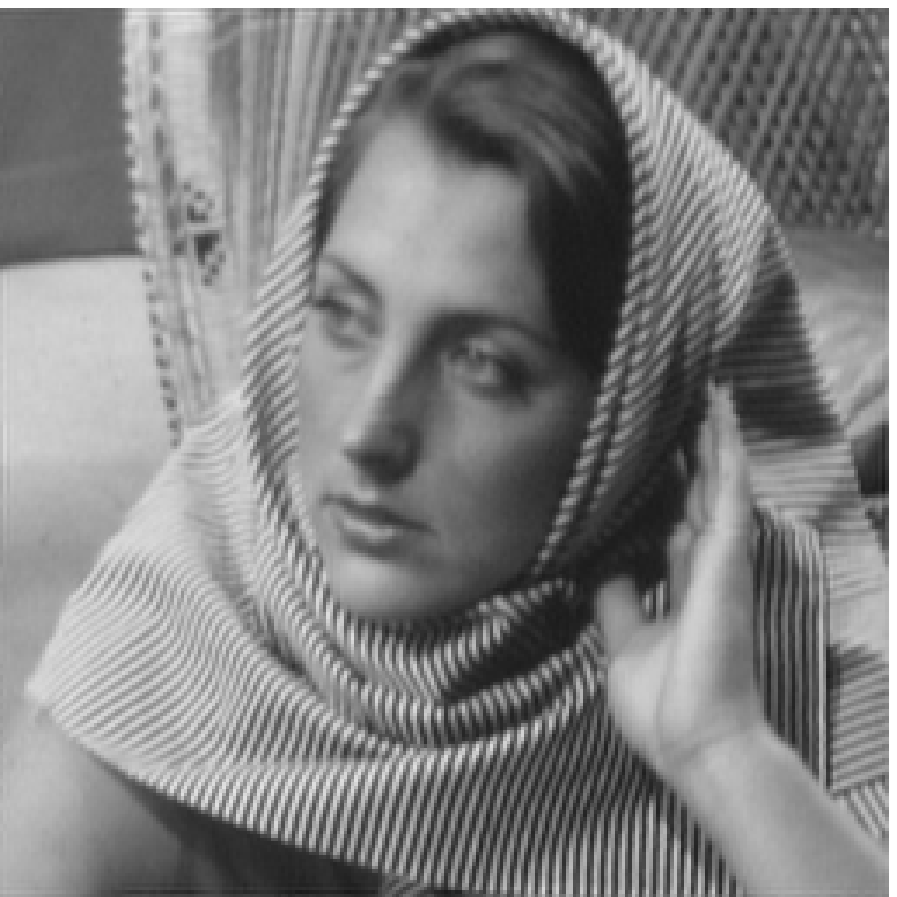} %
\includegraphics[width=0.22\linewidth]{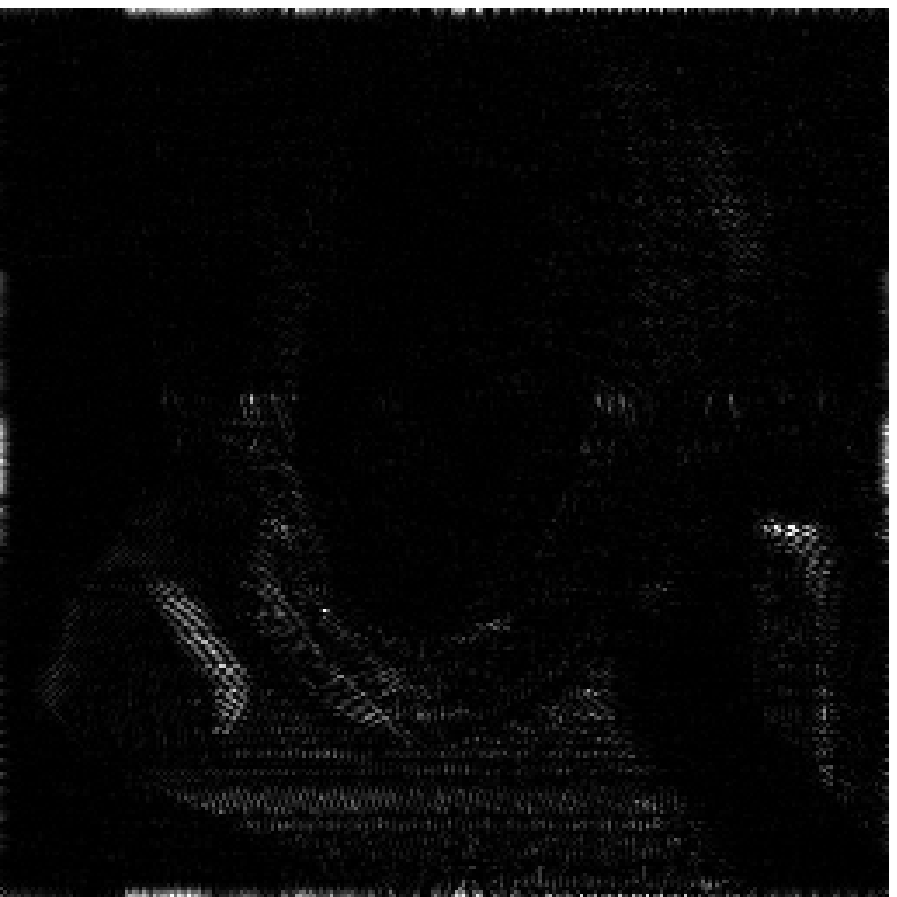} & %
\includegraphics[width=0.22\linewidth]{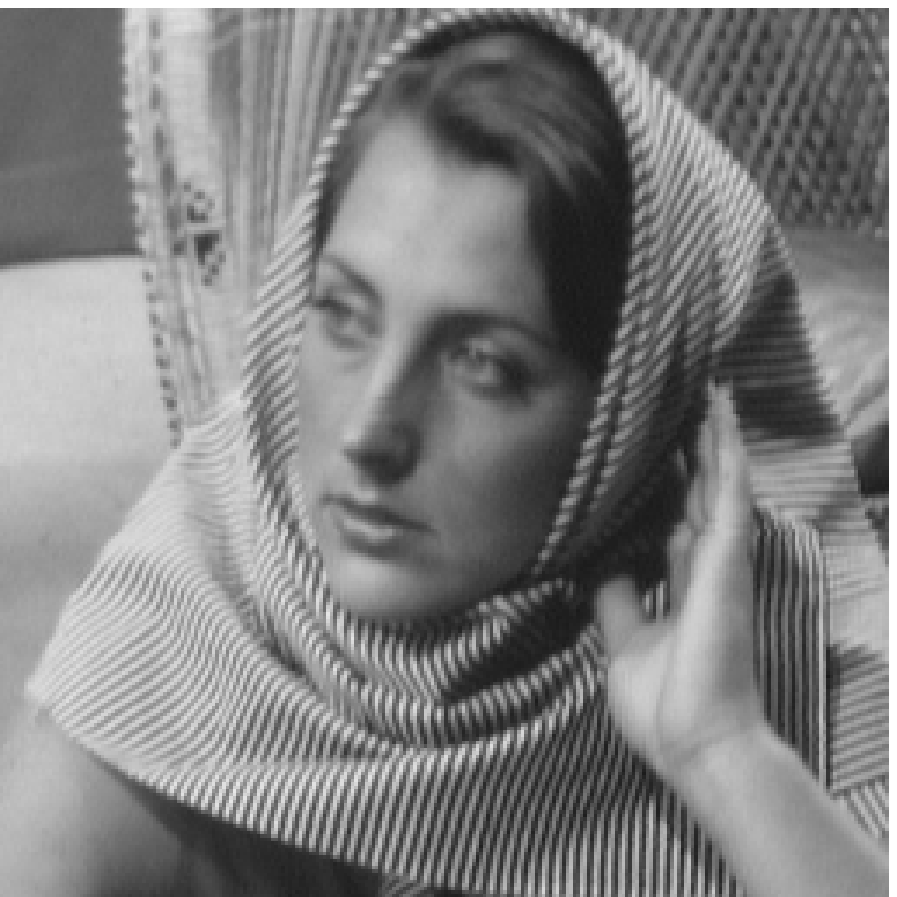} %
\includegraphics[width=0.22\linewidth]{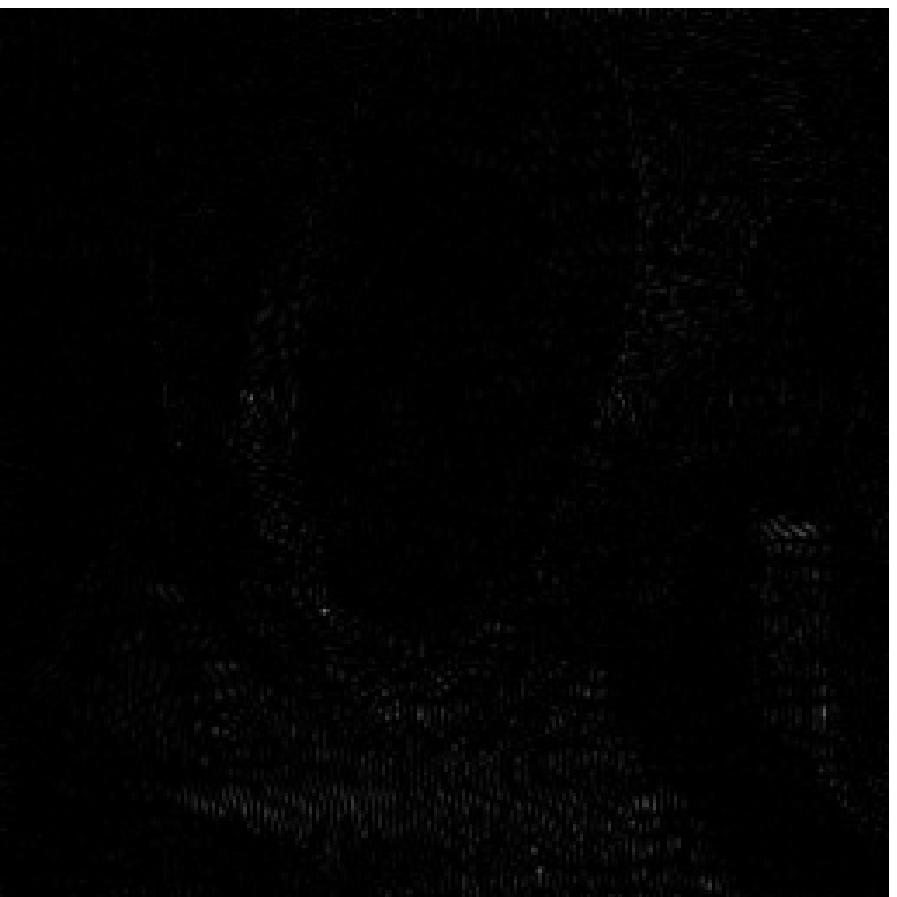} \\
BM3D + CTV, PSNR$=45.45$db & DGD, PSNR$=56.82$db \\
\includegraphics[width=0.22\linewidth]{barbara256BM3Detv00.eps} %
\includegraphics[width=0.22\linewidth]{barbara256BM3Detv00_df.eps} & %
\includegraphics[width=0.22\linewidth]{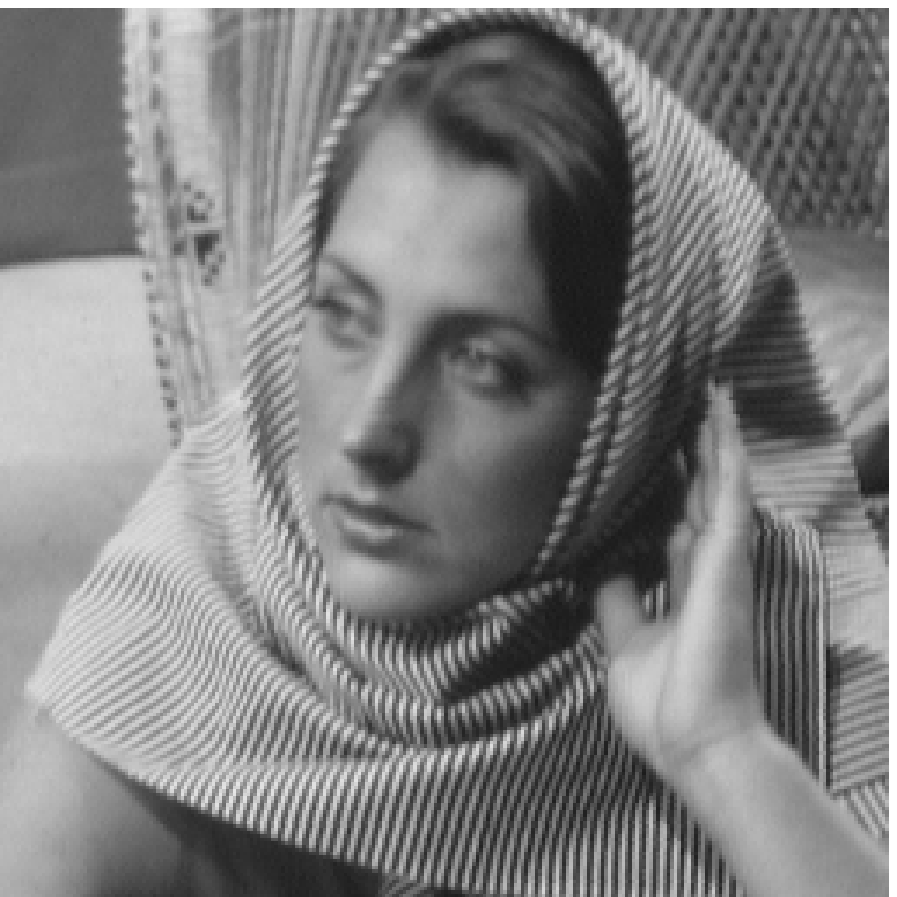} %
\includegraphics[width=0.22\linewidth]{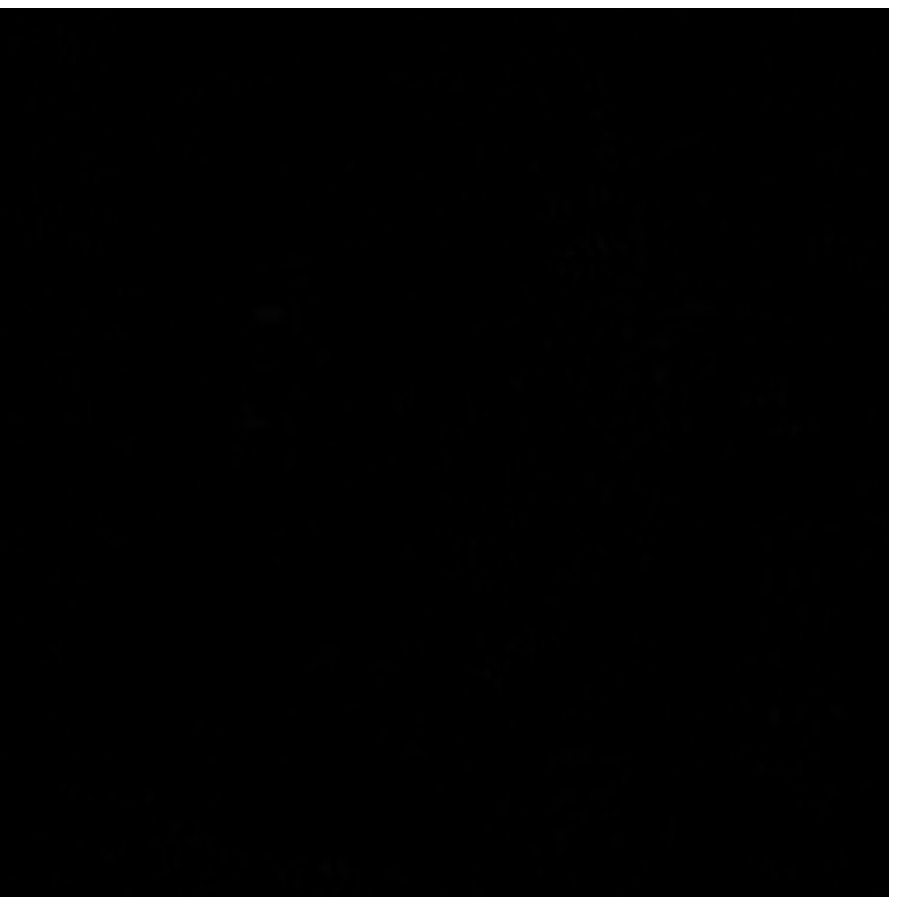}%
\end{tabular}
} \vskip1mm
\par
\rule{0pt}{-0.2pt}%
\par
\vskip1mm 
\end{center}
\caption{{\protect\small Barbara $256\times256$ with Gaussian blur $\protect%
\sigma_b=1$ and Gaussian noise $\protect\sigma_n=0$, the reconstructed image
by different methods and their square errors.}}
\label{Fig Barbara}
\end{figure*}

\begin{figure*}[tbp]
\begin{center}
\renewcommand{\arraystretch}{0.5} \addtolength{\tabcolsep}{-2pt} \vskip3mm {%
\fontsize{8pt}{\baselineskip}\selectfont
\begin{tabular}{cc}
Original and degraded image & Tikhonov  + NL/H$^1$, PSNR$=22.10$db \\
\includegraphics[width=0.22\linewidth]{cameraman.eps} %
\includegraphics[width=0.22\linewidth]{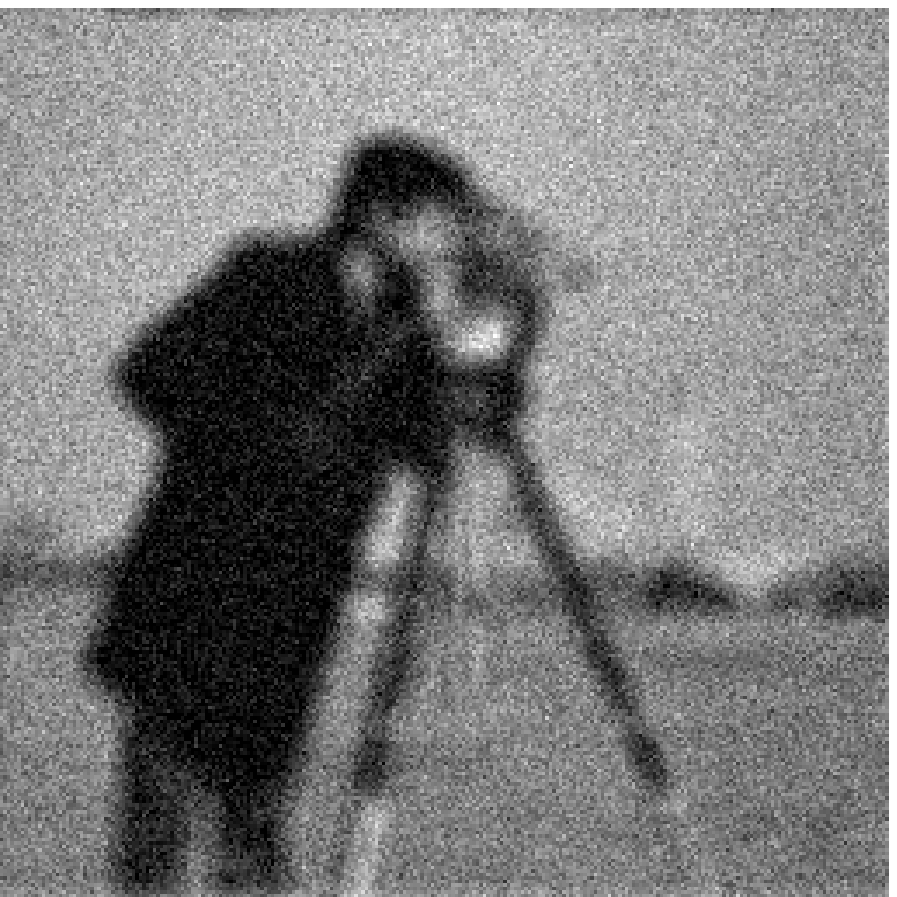} & %
\includegraphics[width=0.22\linewidth]{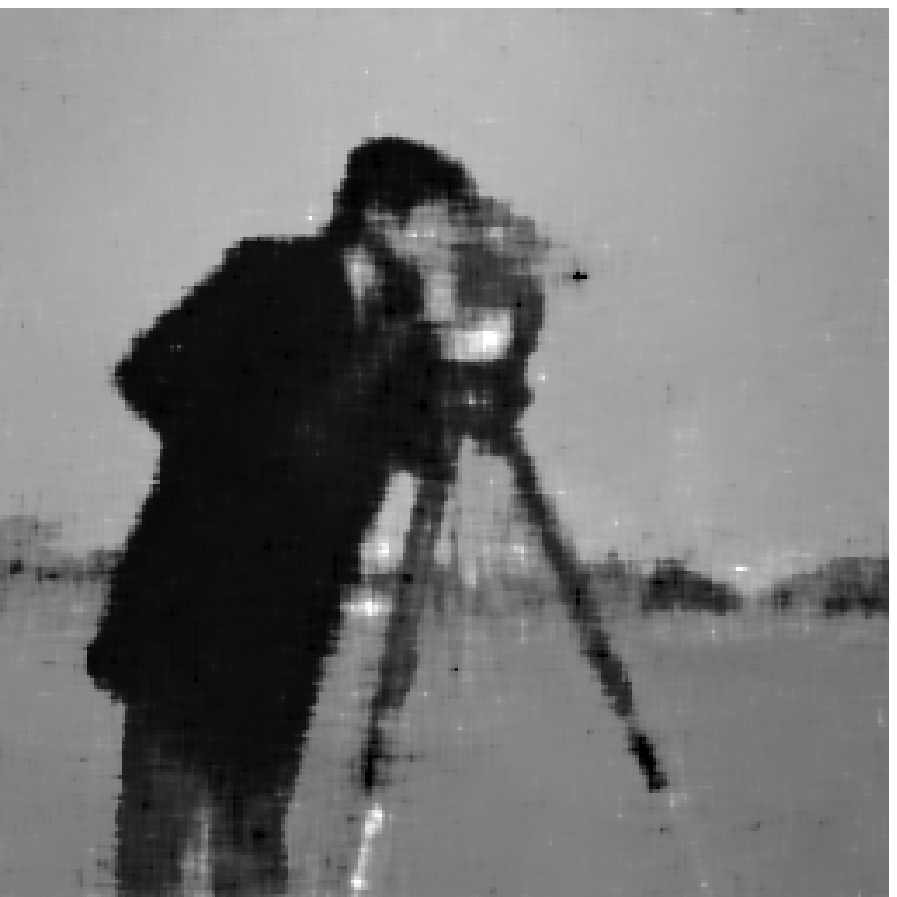} %
\includegraphics[width=0.22\linewidth]{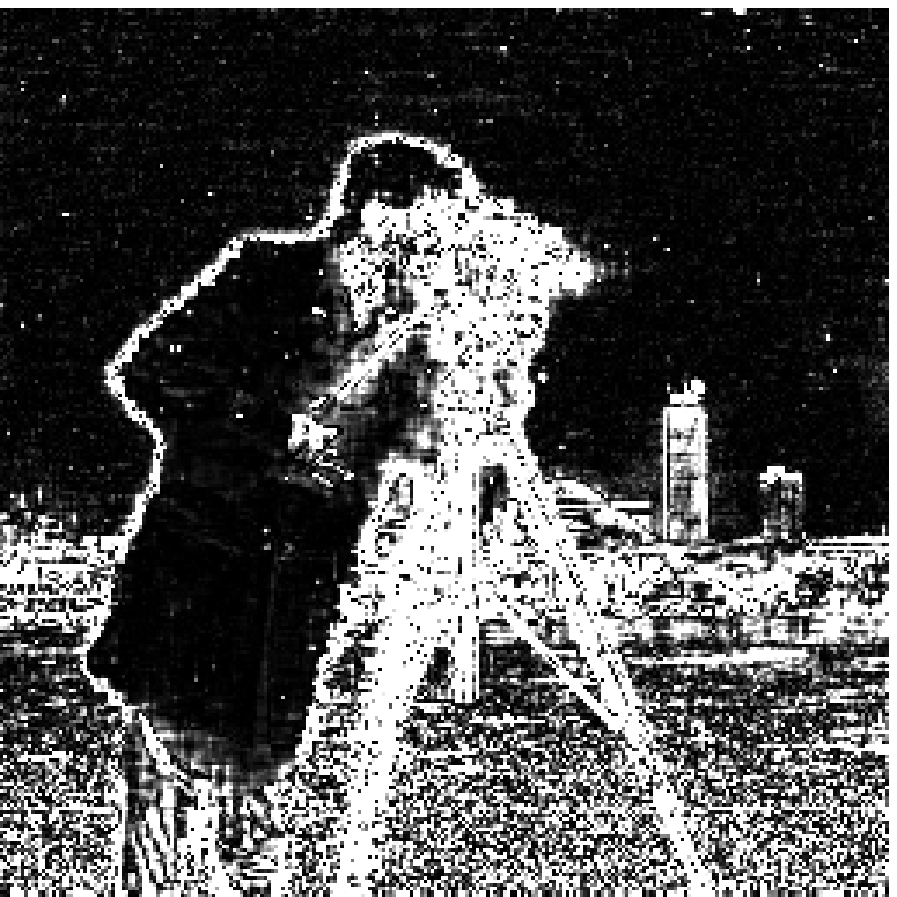} \\
 Tikhonov  +  NL/TV, PSNR$=22.28$db & OPF + CTV, PSNR$=22.58$db \\
\includegraphics[width=0.22\linewidth]{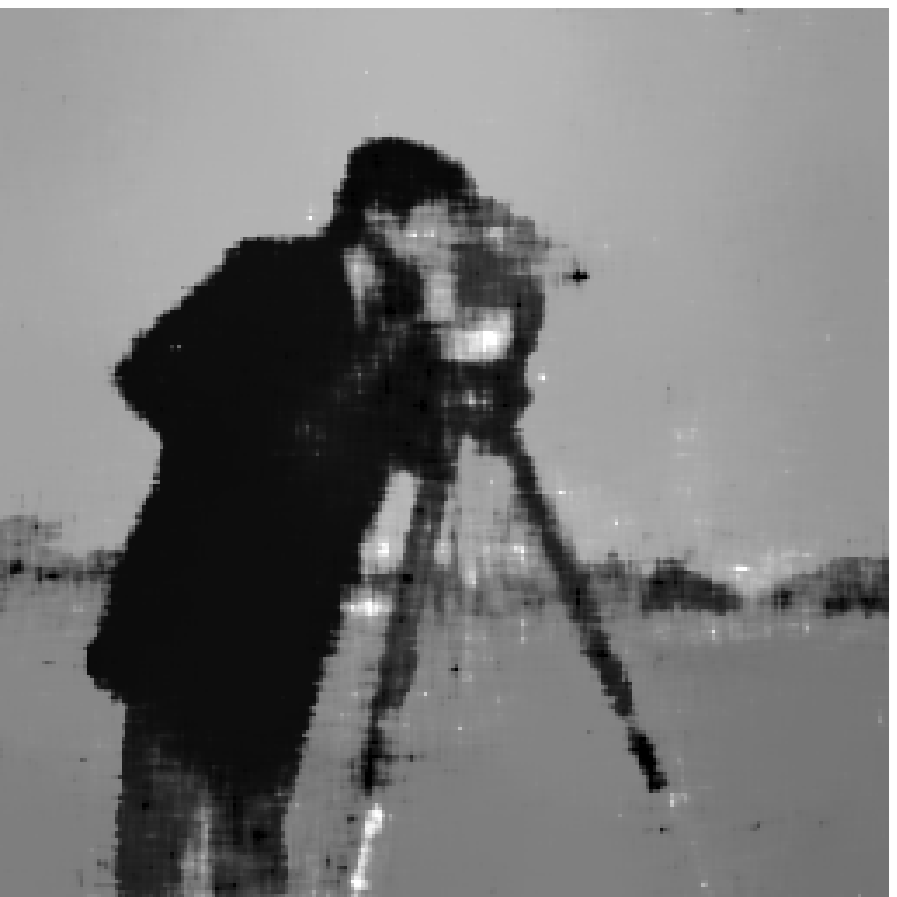} %
\includegraphics[width=0.22\linewidth]{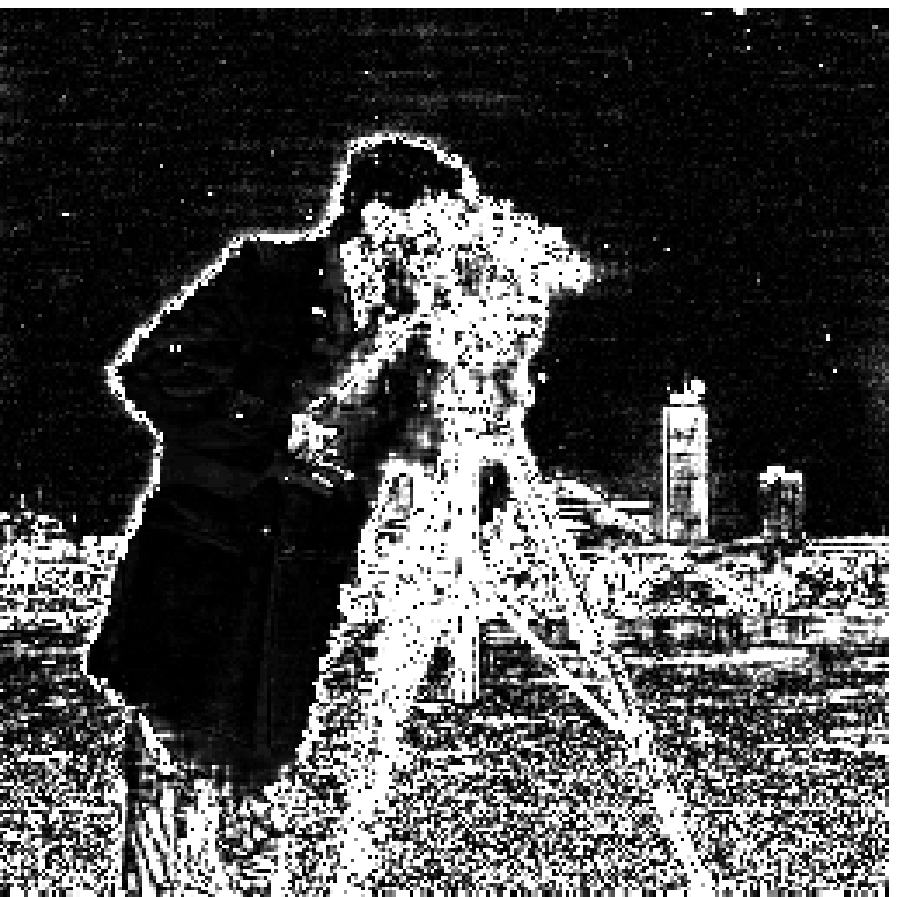} & %
\includegraphics[width=0.22\linewidth]{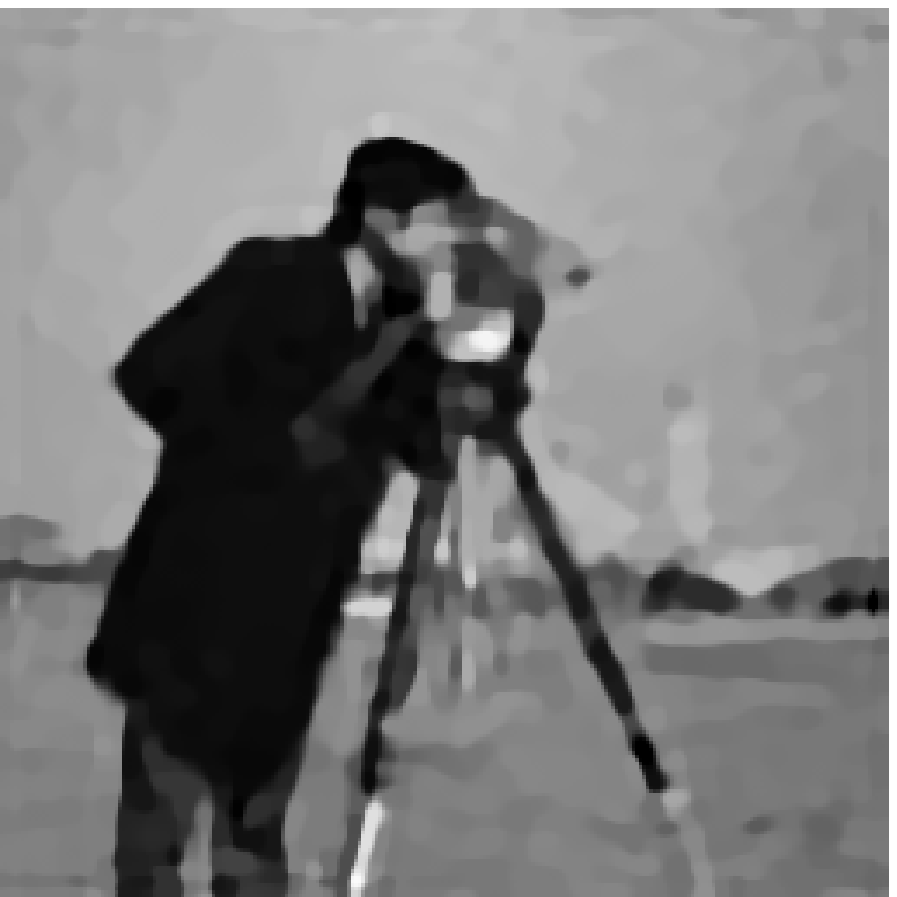} %
\includegraphics[width=0.22\linewidth]{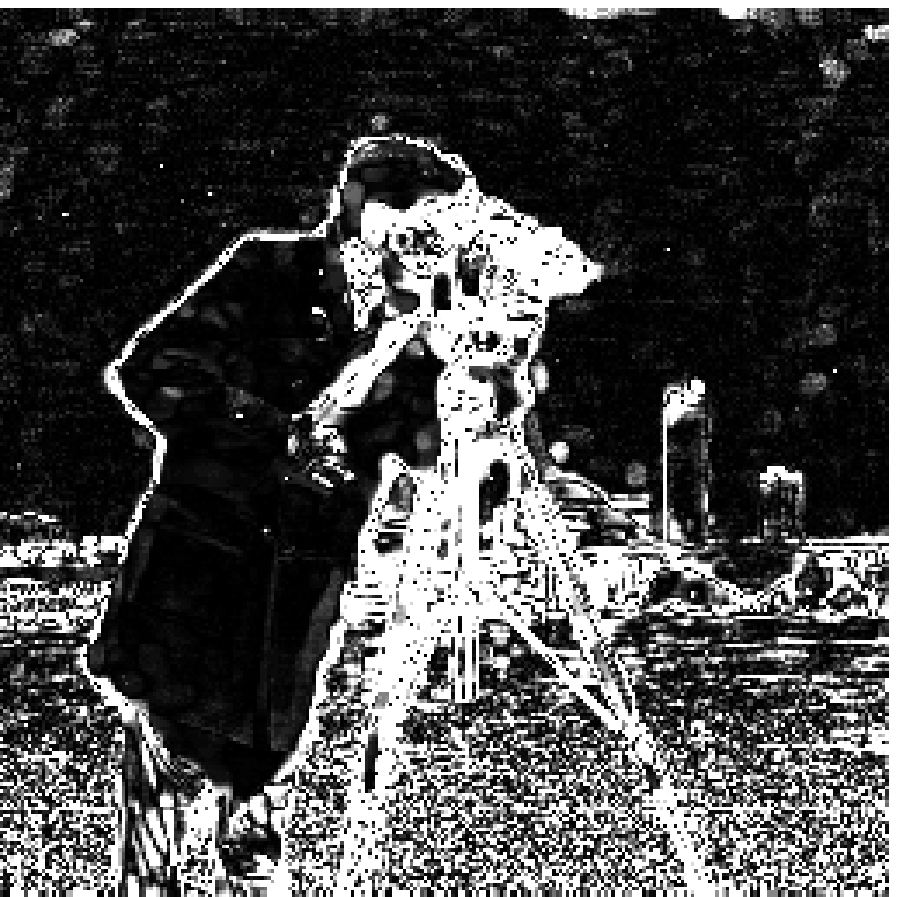} \\
BM3D + CTV, PSNR$=22.80$db &  \\
\includegraphics[width=0.22\linewidth]{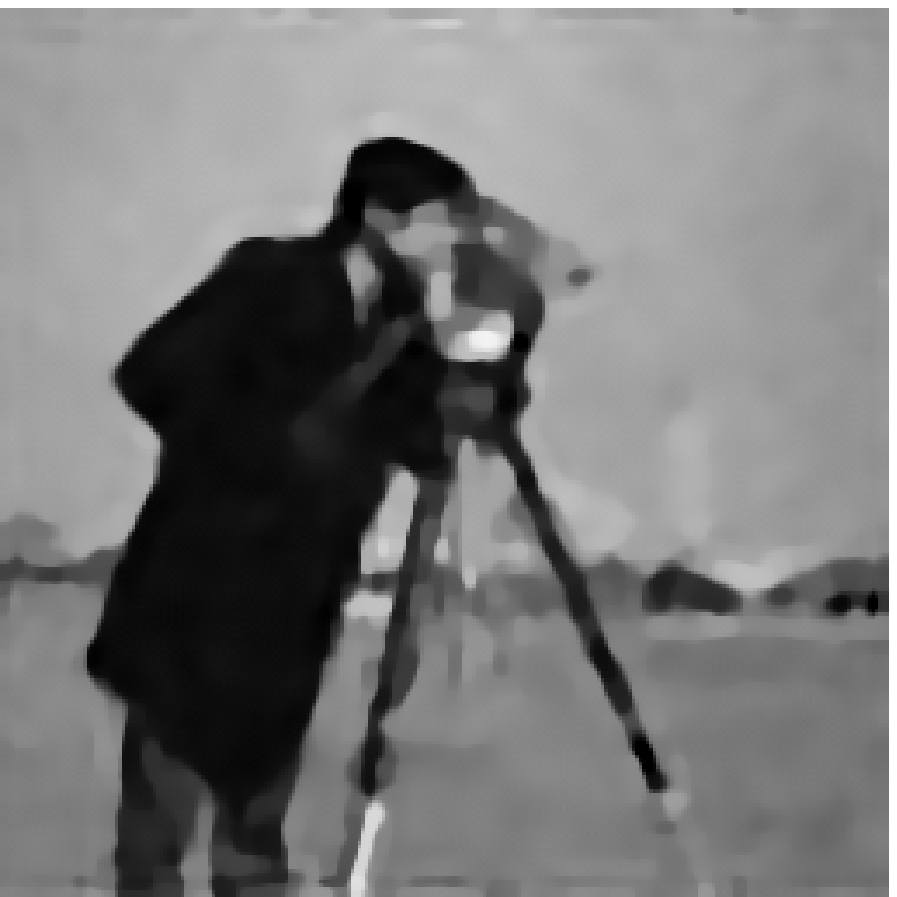} %
\includegraphics[width=0.22\linewidth]{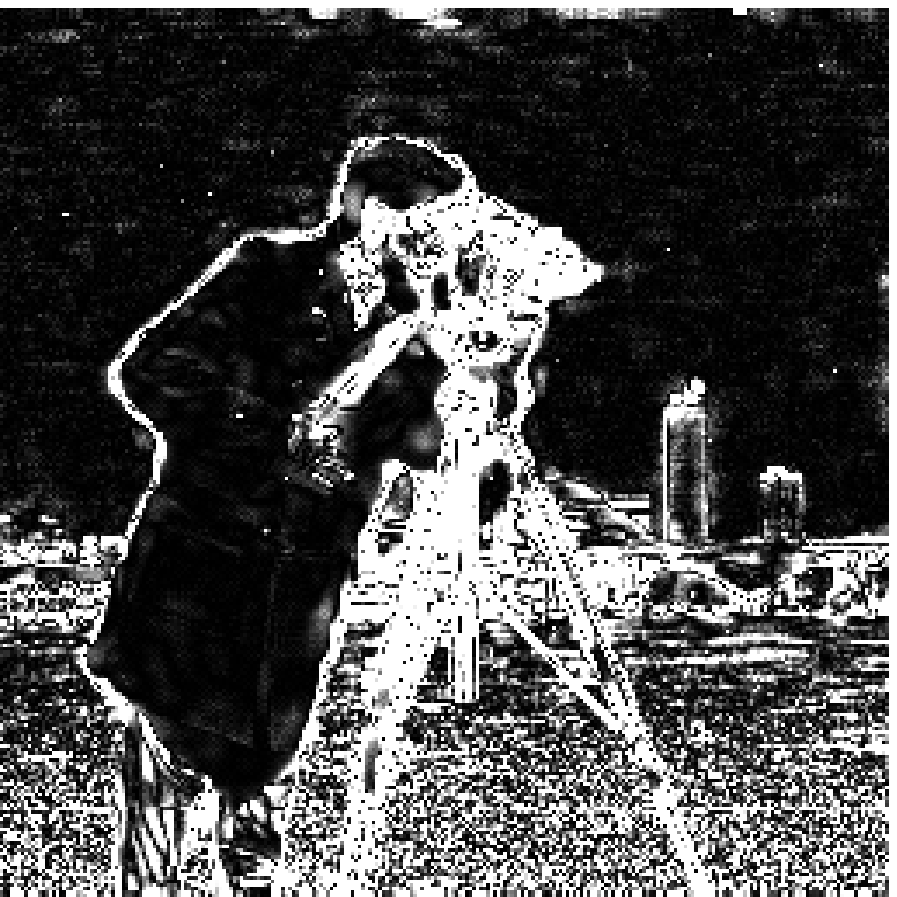} &
\end{tabular}
} \vskip1mm
\par
\rule{0pt}{-0.2pt}%
\par
\vskip1mm 
\end{center}
\caption{{\protect\small Carmeraman $256\times256$ with box average kernel
and Gaussian noise $\protect\sigma_n=20$, the reconstructed image by different methods and their square errors. }}
\label{Fig carmeraman}
\end{figure*}

\section{Conclusion}

The Total Variation (TV) approach is a classical method to reconstruct
images from the noisy or blurred ones. We have reconsidered deeply this
approach and propose a modified version which we call \textit{Controlled
Total Variation }(CTV) regularization. The proposed algorithm is based on a
relaxation of the TV in the classical TV minimization approach which permits
us to find the best $L^{2}$ approximation to the solution of inverse
problems. Numerical simulations show that our CTV algorithm gives superior
restauration results compared to known methods.

\end{document}